\documentclass[journal,10pt]{IEEEtran}
\usepackage{etex}
\pdfoutput=1
\usepackage[table]{xcolor} 
\usepackage{caption}  
\usepackage{colortbl}
\usepackage{diagbox}               
\usepackage{graphicx}
\usepackage{amsmath}
\usepackage[figuresright]{rotating}
\usepackage{supertabular}
\usepackage{threeparttable}
\usepackage{booktabs}
\usepackage{graphicx}
\usepackage{txfonts}
\usepackage{multirow}
\usepackage{verbatim}
\usepackage{array}
\usepackage{subfig}
\usepackage{indentfirst}
\usepackage{pgfplots}
\usepackage{makecell}
\usepackage[colorlinks, linkcolor=black, urlcolor = black, citecolor=black]{hyperref}
\hypersetup{colorlinks=true,citecolor = blue}

\usepackage{tikz}
\RequirePackage{amsmath,mathrsfs,amsfonts,amssymb}
\usepackage{multirow}  
\newtheorem{definition}{Definition}   

\usepackage{amsmath}
\usepackage{arydshln} 
\usepackage{color} 
\usepackage{algorithm}
\usepackage{algorithmic}

\begin{document}

\title{Template for IEEE TVT \LaTeX\ Submission}

\title{VSRQ: Quantitative Assessment Method for Safety Risk of Vehicle Intelligent Connected System}

\author{
	
	Tian Zhang,
	Wenshan Guan,
	Hao Miao, 
	Xiujie Huang,~\IEEEmembership{$Member,~IEEE$,}
	Zhiquan Liu,~\IEEEmembership{$Member,~IEEE$,}
	Chaonan Wang,~\IEEEmembership{$Member,~IEEE$,}
	Quanlong Guan*\thanks{*Corresponding author},~\IEEEmembership{$Member,~IEEE$,}
	Liangda Fang,~\IEEEmembership{$Member,~IEEE$,}
	Zhifei Duan       

	\IEEEcompsocitemizethanks{
		
		\IEEEcompsocthanksitem Tian Zhang and Wenshan Guan are with the Department of Cyberspace Security, College of Information Science and Technology, Jinan University, Guangzhou 511486, China (e-mail: tianzhang@stu2022.jnu.edu.cn; ggwsh@stu2022.jnu.edu.cn).
		\IEEEcompsocthanksitem Hao Miao is with the Department of Computer Science, College of Information Science and Technology, Jinan University, Guangzhou 510632, China (e-mail: miaohao77@stu2020.jnu.edu.cn).
		\IEEEcompsocthanksitem Xiujie Huang is with the College of Information Science and Technology, Jinan University, Guangdong Institution of Smart Education, Jinan University, Guangzhou 510632, China (e-mail: t{\_}xiujie@jnu.edu.cn).
		\IEEEcompsocthanksitem Zhiquan Liu is with the College of Cyber Security, Jinan University, Guangzhou 510632, China. (e-mail: zqliu@vip.qq.com)	
		\IEEEcompsocthanksitem Chaonan Wang is with the College of Information Science and Technology, Jinan University, Guangzhou 510632, China, also with the Guangdong Gene Data Processing and Analysis Engineering Research Center, China (e-mail: c{\_}wang@jnu.edu.cn).
		\IEEEcompsocthanksitem Quanlong Guan is with the Department of Computer Science, College of Information Science and Technology, Jinan University, Guangzhou 510632, China (e-mail: gql@jnu.edu.cn). 	
		\IEEEcompsocthanksitem Liangda Fang is with the Department of Computer Science, College of Information Science and Technology, Jinan University, Guangzhou 510632, China, also with the Pazhou Laboratory, Guangzhou 510330, and also with the Guangxi Key Laboratory of Trusted Software, Guilin University of Electronic Technology, Guilin 541004, China (e-mail: fangld@jnu.edu.cn).
		\IEEEcompsocthanksitem Zhifei Duan is with the Guangzhou XPeng Motors Technology Co., Ltd No.8 Songgang Road, Changxing Street, Cencun, Tianhe District, Guangzhou, China (e-mail: duanzf@xiaopeng.com).	
	}

}

{}
\maketitle

\begin{abstract}
The field of intelligent connected in modern vehicles continues to expand, and the functions of vehicles become more and more complex with the development of the times.
This has also led to an increasing number of vehicle vulnerabilities and many safety issues.
Therefore, it is particularly important to identify high-risk vehicle intelligent connected systems, because it can inform security personnel which systems are most vulnerable to attacks, allowing them to conduct more thorough inspections and tests.
In this paper, we develop a new model for vehicle risk assessment by combining I-FAHP with FCA clustering: VSRQ model. 
We extract important indicators related to vehicle safety, use fuzzy cluster analys (FCA) combined with fuzzy analytic hierarchy process (FAHP) to mine the vulnerable components of the vehicle intelligent connected system, and conduct priority testing on vulnerable components to reduce risks and ensure vehicle safety.
We evaluate the model on OpenPilot and experimentally demonstrate the effectiveness of the VSRQ model in identifying the safety of vehicle intelligent connected systems. 
The experiment fully complies with ISO 26262 and ISO/SAE 21434 standards, and our model has a higher accuracy rate than other models. 
These results provide a promising new research direction for predicting the security risks of vehicle intelligent connected systems and provide typical application tasks for VSRQ. 
The experimental results show that the accuracy rate is 94.36{\%}, and the recall rate is 73.43{\%}, which is at least 14.63{\%} higher than all other known indicators.
\end{abstract}

\begin{IEEEkeywords}
	vehicle safety, ISO 26262, ISO/SAE 21434, FCA, FAHP, intelligent connected system, vulnerable components  
\end{IEEEkeywords}

\IEEEpeerreviewmaketitle

\section{Introduction}

\IEEEPARstart{V}{ehicle} safety is the study and practice of design, construction, equipment and regulations to minimize the occurrence and consequences of traffic accidents involving motor vehicles.
When the vehicle was first invented in 1886, there were no safety components.
When people buy a vehicle, they mainly pay attention to the physical characteristics of the vehicle.
Then with the development of vehicle, seat belts appeared in vehicles in 1959, child seats were generally equipped on vehicles in 1970, and airbags were generally installed in vehicles in 1980.
The electronic application of vehicle safety is marked by the emergence of ABS (Anti-lock Braking System) invented by Bosch in Germany in the 1950s.
Since the 21st century, with the development of vehicle technology and the continuous improvement of people's requirements for vehicle safety performance, vehicle safety has become particularly important.
It is necessary to predict and assess the risk of vehicle.

So far, there are many models for predicting and evaluating the risks of vehicle, such as EVITA\cite{henniger2011evita}, "automotive functional safety = safety + security"\cite{hommes2012review}, HEAVENS 1.0\cite{IslLS2016}, AVSDA\cite{moukahal2022avsda}, RLE\cite{ying2021reputation}, P2BA\cite{feng2021p2ba}, TROVE\cite{guo2020trove}, etc.
Although these models can accurately quantify the safety of vehicles, the quantitative results of these models are all a fixed value. In fact, due to the influence of various complex factors such as materials and temperature, the results of safety quantification are more accurate within the interval.
However, there are still some challenges in creating a good quantitative model for vehicle risk assessment: 
First, there is no model to predict and evaluate the risk of the vehicle before landing at present, and due to the independence of vehicle parts, the risk prediction before the landing of the vehicle is more uncertain than after the landing.
In addition, because the data before the vehicle lands is not public, it is privatized by various vehicle companies.

In order to solve the above challenges, predecessors have proposed some models and made some risk projections over the last three years. Using the attack surface of automated driving for sensors, operating systems, control systems, and internet of vehicles connections as an example, \cite{gao2021autonomous} and \cite{de2020driverless} examined the security of autonomous driving. 
\cite{zelle2022threatsurf} based on automatic threat surface assessment as an automatic identification method in automotive network security engineering for difficulty analysis.
\cite{behfarnia2018risk} evaluated the risk associated with autonomous vehicles using a Bayesian defense map.
\cite{cui2020edge} proposed an efficient and privacy-preserving VANET data download scheme based on the concept of edge computing.
\cite{kong2021blockchain} proposed mechanism first achieves the verifiable aggregation and immutable dissemination of performance records by exploiting a blockchain with the proof-of-stake (PoS) consensus. 
In this paper, we propose a model named VSRQ. 
The model standard is ISO 26262\cite{PalWH2011} and ISO/SAE 21434\cite{MacSV2020} and we break its risk levels into serveral tiers.
The vehicle intelligent connected system is seriously at risk when the VSR is larger than or equal to 0.758, the vehicle is in a dangerous condition, and the system needs to be rectified right away.
The vehicle intelligent connected system is in a mild and severe risk condition, and the vehicle state is in a critical state, when VSR is greater than or equal to 0.506 and less than 0.758.
At this point, a detailed study should be carried out based on the first stage's vulnerable components.
If the first-stage influencing factors' Electronic Control Unit (ECU) coupling risk is at a high level, the entire vehicle's ECU system needs to be corrected until the risk is normal or minimal.
vehicle intelligent connected system risk is normal or minimal and the vehicle state is steady state when VSR is more than or equal to 0 and less than 0.506.
There is no need to update the vehicle intelligent connected system\cite{PalWH2011} at this moment because it has attained the quality management level.
Finally, we also proposed a vehicle software risk assessment system based on the VSRQ model, using the FCA algorithm to adjust the weight of vehicle indicators, setting dynamic parameters, and using the pure puresuit (PP) and FAHP algorithms to establish a vehicle intelligent connected risk quantitative assessment system, which is convenient for software engineers to advance the purpose of predicting the vulnerable components of the vehicle is to check the possible risks of the vehicle in time before the vehicle lands, so as to avoid potential dangers.
Compared with other models, VSRQ has the following advantages:  
First, VSRQ can predict the risk of vehicles within a reasonable range, solving the problem that the previous models were too absolute in the prediction of vehicle safety risk parameters. 
Second, our model also analyzes vulnerable components, automatically and effectively identifies potentially vulnerable components in the vehicle, calculates the security level of on-board software, and systematically discovers the security of on-board integrated systems. 

The remaining chapters of this paper are organized as follows: the Section \ref{sec:relatedwork} briefly reviews some of the previous work related to vehicle safety. Metric definitions and problem statements for the safety of vehicle intelligent connected systems are given in the Section \ref{sec:preliminaries}. 
The details of the proposed VSRQ model are described in the Section \ref{sec:algorithm}. 
An evaluation of the effectiveness and efficiency of the model is provided in the Section \ref{sec:experiments}. 
Finally, the full text is summarized and conclusions are given in the Section \ref{sec:conclusion}.

\section{Related Work}
\label{sec:relatedwork}
The concept of integration is a concept derived from the increase in the complexity of the vehicle control unit, and is defined by the original IT industry standards. 
The safety assessment of the vehicle intelligent connected system is a relatively difficult task, it is quite different from the conventional software system.
It does not have a central processing system, but uses the ECU interaction on the vehicle to achieve the function of the vehicle, and the communication means and technology of the vehicle are also quite different from other systems \cite{2008Vehicle}. 
\begin{figure}
	\centering
	\includegraphics[scale=0.27]{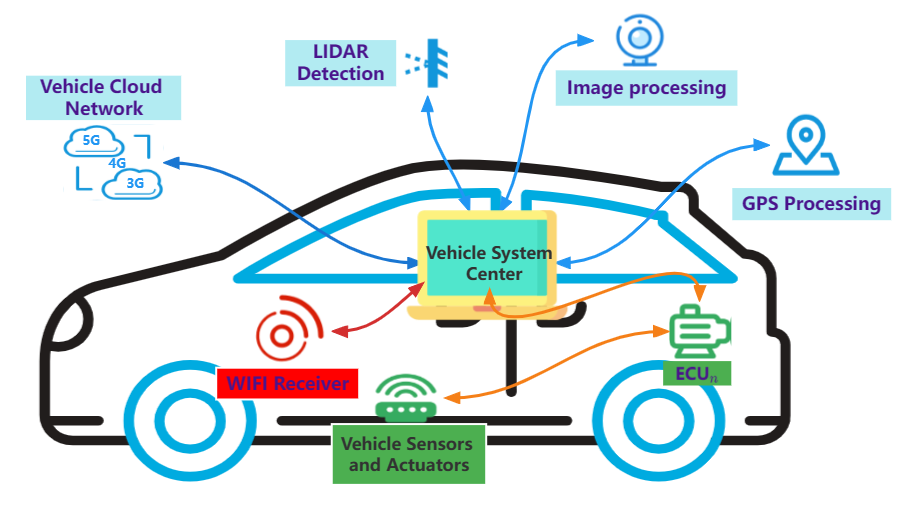}
	\caption{A General vehicle intelligent connected system Model}
	\label{fig:car2}
\end{figure}
\subsection{Development Of International Standards}

Existing vehicle safety standards are evaluated based on international standards, including the  “ISO 26262 Road vehicles--Functional safety” standard published in 2011\cite{PalWH2011}. 
Then came the "SAE J3601 Cybersecurity Guidebook for Cyber-Physical Vehicle Systems"\cite{sae2016cybersecurity} released in 2016, which defines a structured cybersecurity process framework in detail, which means that for the first time automotive information security will be elevated to an equal or even more important position than functional safety.
Finally, in August 2021, ISO and SAE released the first international standard in the field of automotive information security, ISO/SAE 21434 Road Vehicles - Cybersecurity Engineering \cite{MacSV2020}, which specifies the detailed risk assessment process for automotive development projects, and proposes the vehicle information security risk assessment method - TARA. 
TARA analysis includes steps such as asset identification, threat scenario identification, attack path analysis, attack feasibility rating, risk level assessment, and risk handling measures, but the standard does not specify specific implementation methods.

\begin{figure*}
	\centering
	\includegraphics[width=17cm,height=7cm]{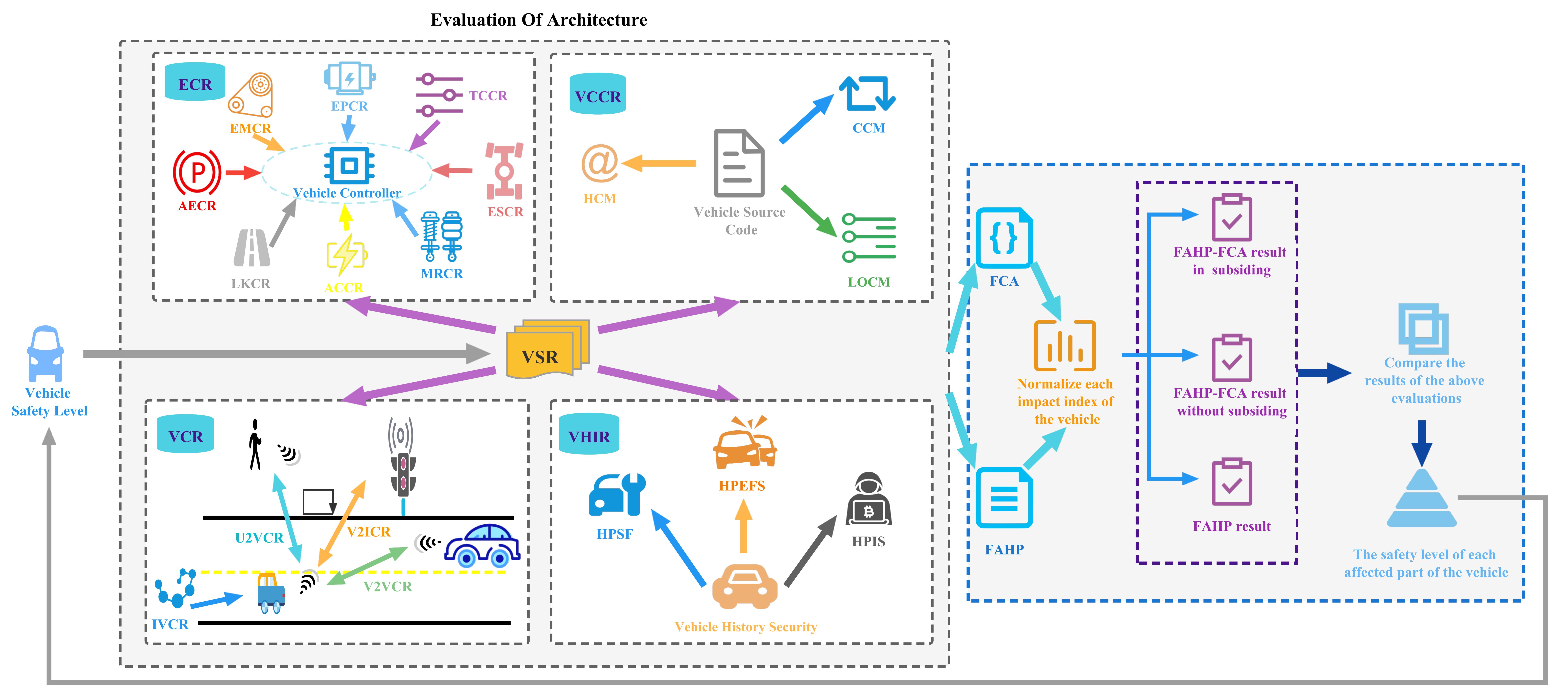}
	\caption{Vehicle risk analysis structure diagram}
	\label{fig:structure1}
\end{figure*}
\subsection{Existing Model Framework Development}

The EVITA Threat and Risk Model \cite{henniger2011evita} was first published in 2009 and is considered to be one of the effective risk assessment models for the automotive industry and a pioneering risk rating method for automotive E/E systems. 
The model focuses on identifying all possible attacks against a specific target, taking into account the four parameters of risk rating, security, financial, operational, and privacy, but the model does not study the proportions of the four parameters in sufficient detail.
Then Burton et al. expanded the ISO 26262 standard in 2012 and proposed a comprehensive approach of "automotive functional safety = safety + security" \cite{hommes2012review}.
They analyzed the third hazard of deliberate manipulation of electrical control systems through the open interface of the system, providing better protection and safety for the vehicle. 
Then Wolf and Scheibel et al. \cite{BurLV2012}integrated existing techniques into a risk-rating framework for automotive systems and used perasset to define attack trees based on safety questionnaires.
Following the implementation risk assessment framework HEAVENS1.0 \cite{IslLS2016} proposed by Islam et al., the framework is fully consistent with the functional safety standard ISO 26262, and introduces the "safety level" based on the ASIL concept in ISO 26262.
Finally, Moukahal et al. proposed the AVSDA safety decay evaluation framework \cite{moukahal2022avsda} in 2022, which can quantitatively evaluate automotive software systems, and is not limited by the development stage, and can evaluate the safety of the vehicle in the operating stage.
\section{Preliminary}
\label{sec:preliminaries}
In this section, we first define the vulnerability assessment structure for vehicle systems, and then define the impact of various factors on the severity of vehicle damage.
According to the general vehicle intelligent connected system model shown in Fig. \ref{fig:car2}, we designed the vehicle risk analysis structure diagram proposed in this paper as shown in Fig. \ref{fig:structure1}.


\subsection{Vulnerable Components Assessment Structure}


We adopt a second-order multi-index system to quantify the intelligent connected system risk in the concept stage of the vehicle. 
Fig. \ref{fig:structure1} shows our process for evaluating vehicle software systems. 
It consists of two parts, the first part is to quantitatively evaluate the vulnerable parts of the vehicle system, and the second part is to evaluate the risk level of the vehicle system and determine the possibility of the vehicle system being attacked. 

\begin{figure*}[!]
	\centering
	\includegraphics[width=15.5cm,height=9.98cm]{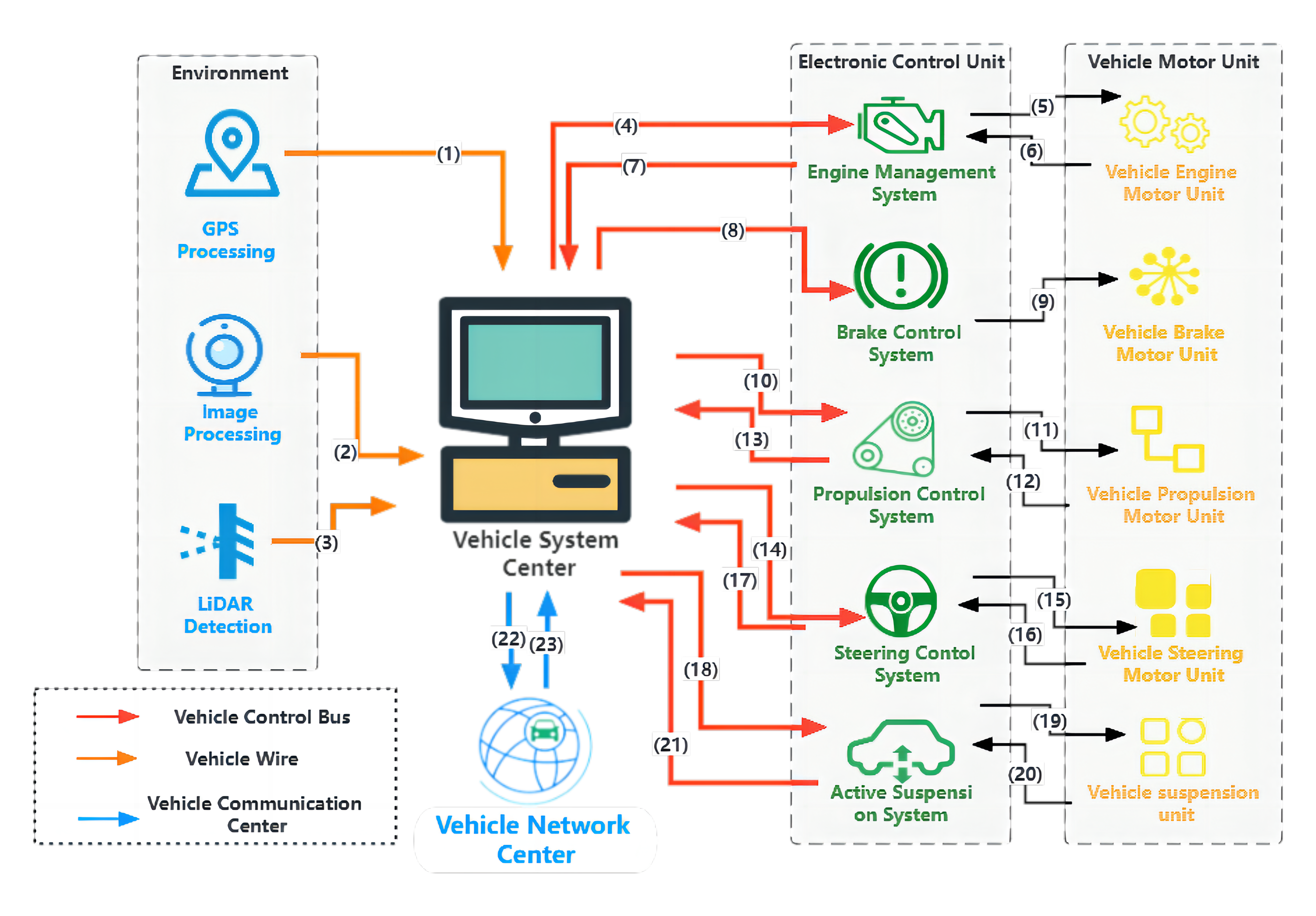}
	\caption{Example diagram of information flow in vehicle intelligent connected system}
	\label{fig:Pro}
\end{figure*}

First of all, in order to analyze the  influencing factors of the vulnerable components of the vehicle intelligent connected system, all information flows related to the vehicle software system should be considered, as shown in Fig. \ref{fig:Pro}. Fig. \ref{fig:Pro} describes the information flow in Fig. \ref{fig:car2}, where Vehicle Communication Center represents the Internet of Vehicles. 
The following information is provided: (1)GPS Processing (vehicle location); (2)Image Processing(vehicle monitoring and tracking); (3)LIDAR Detection(surrounding environment monitoring); (4)(5)vehicle start or stop command; (6)(7)vehicle startup or shutdown status data; (8)(9)brake control command(such as vehicle brake position); (10)(11)vehicle propulsion control command(such as propulsion target distance); (12)(13)vehicle propulsion status data(such as actual advance position); (14)(15)vehicle steering control instruction(such as target steering angle); (16)(17)vehicle steering status data(such as actual steering angle); (18)(19)control vehicle smooth driving command(such as vehicle turning spring deformation); (20)(21)adjust vehicle suspension height(such as reduce body position changes); (22)actual state data of the vehicle(such as surroundings of the vehicle, vehicle system hazard); (23)vehicle network center data(such as transfer vehicle information). 
It's important to note that, in accordance with \cite{bertolino2018tour,francia2021automotive}, we divide the communication risk (VCR) of the vehicle system into four categories based on the vehicle's four communication domains. These categories are: in-vehicle communication risk (IVCR), user to vehicle-to-vehicle communication risk (U2VCR), vehicle to vehicle communication risk (V2VCR), and vehicle to infrastructure communication risk (V2ICR).

Then according to the information flow model in Fig. \ref{fig:Pro}, our group divides the evaluation of the vulnerable component structure of vehicle systems into the following three levels.
The first level is the target levels (the vulnerable components of the vehicle system).
The second level is the indicator layer, including the coupling risk of the vehicle components(ECU Coupling), the communication risk, the complexity risks, and history of security issues.
The third level consists of sub-indicator layers of the second level, as shown in Fig. \ref{fig:structure1}.
In the following, we will define the indicators of each layer in detail.
\subsubsection{ECU Coupling Risk (ECR)}
ECU (Electronic Control Unit) electronic control unit, like ordinary computers, it consists of microcontroller (MCU), memory (ROM, RAM), input/output interface (I/O), analog-to-digital converter (A/D), and large-scale integrated circuits such as shaping and driving.

In theory, the more complex the function, the greater the coupling of the ECU, while the ECU coupling allows malicious messages to propagate, which also leads to a decrease in the safety performance of the vehicle as the ECU coupling increases.
Assuming that a malicious message is propagated in a vehicle ECU with complex components, and the malicious message can be propagated from this ECU to another ECU, where more entities in the vehicle are connected by control, this behavior will cause extreme damage to the vehicle. big damage.
Therefore, the degree of coupling of ECUs should be considered when identifying vulnerable components in vehicle software systems.

In the coupling of this ECU, we do not consider that the network is interconnected through the bus topology, but will analyze the communication risk of the bus topology in the second-order indicators, and their specific introduction is as follows:

\begin{definition}[Engine Management System Coupling Risk (EMCR)]
	\rm By controlling air intake, fuel injection, and ignition to achieve the balance of engine power, economy, emissions and other performance, the torque analysis function of the whole vehicle is integrated in Engine Management System(EMS).
	

\end{definition}

\begin{definition}[Transmission Control Unit Coupling Risk (TCCR)] 
	\rm Control the oil pressure through the solenoid valve, realize the automatic engagement or separation of the clutch, complete the gear switch at the right time, and improve the power and smoothness of the vehicle. 
\end{definition}

\begin{definition}[Electric Power Steering Coupling Risk (EPCR)] 
	\rm The motor assists the driver in steering, reducing the difficulty of driving. 
\end{definition}

\begin{definition}[Body Stability Control Coupling Risk (ESCR)] 
	\rm It integrates functions such as TCS, ABS, ESC, etc., and realizes the stable driving of the vehicle by controlling the braking force at the wheel end. 
\end{definition}

\begin{definition}[Active Suspension System Coupling Risk (MRCR)] 
	\rm Control the solenoid valve to adjust the height or damping of the suspension system to improve the driving stability and comfort of the vehicle.
\end{definition}
\begin{definition}[Adaptive Battery Life Coupling Risk (ACCR)] 
	\rm The vehicle follows the driver's desired speed as much as possible by controlling the driving force or braking force while maintaining a safe driving distance according to the movement of the preceding vehicle, freeing the driver's hands and feet. 
\end{definition}
\begin{definition}[Lane Keeping Assist Coupling Risk (LKCR)] 
	\rm On the basis of identifying the lane lines, control the steering wheel angle to keep the vehicle in the center of the lane.
\end{definition}
\begin{definition}[Automatic Emergency Braking Coupling Risk (AECR)] 
	\rm In an emergency, control the braking force to stop the vehicle to avoid a collision accident. 
\end{definition}

In the category of ECU coupling risk, the coupling equation of each subsystem is defined as follows:
\begin{equation}\label{key}
	E = \sum _{i=1, j=1}^{N} \delta_{ij} /2
\end{equation}
where $N$ represents the total number of sub-parts in the system in the vehicle. 
When there is a connection between the two subsystems $s_i$ and $s_j$, it means that there is a calling relationship between them, then the value of $\delta_{ij}$ is 1, otherwise the value is 0.

Taking EMSC as an example, where $N$ represents the total number of sub-parts in the Engine Management System Coupling(EMSC) system in the vehicle. 
When there is a connection between the two subsystems EMSC$_i$ and EMSC$_j$, the value of $\delta_{ij}$ is 1, otherwise the value is 0.

%

\subsubsection{Vehicle Communication Risk (VCR)}

Vehicle information and communication systems are systems that give drivers the most recent, essential information about the state of the road. 
We divide the communication risks of vehicle systems into four categories based on four communication domains of vehicles.
\begin{definition}[In-Vehicle Communication Risks (IVCR)] 
	\rm In-vehicle communication refers to the only non-external communication in the connected vehicle, which satisfies all communication between different ECUs. 
	In-vehicle communication uses networks such as LIN (Local Interconnect Network), CAN (Controller Area Network), FlexRay and MOST (Media Oriented System Transport) \cite{bertolino2018tour}. 
	Therefore, our in-vehicle communication risk will be defined as
	\begin{equation}\label{key2}
		V_1 = \sum _{i=1}^{N} \varrho_1 n_{LIN_{i}} + \varrho_2 n_{CANL_{i}} + \varrho_3 n_{CANH_{i}} + \varrho_4 n_{FL_{i}}  + \varrho_5 n_{MOST_{i}}
	\end{equation}
	$\varrho$ represents the severity of each communication attack on the vehicle, $n_{LIN}$ represents the number of modules in the vehicle that use the LIN network for communication, $n_{CANH}$ represents the number of modules in the vehicle that use the high-speed CAN network for communication, and $n_{CANL}$ represents is the number of modules in the vehicle using the low-speed CAN network for communication, $n_{FL}$ represents the number of modules in the vehicle using the FlexRay network for communication, and $n_{MOST}$ represents the number of modules in the vehicle using the MOST network for communication, and $i$ represents each module in the vehicle, and $N$ represents the total system module of the vehicle.
	
\end{definition}
\begin{definition}[User-To-Vehicle Communication Risks (U2VCR)] 
	\rm User-To-Vehicle(U2V) mainly refers to providing in-vehicle services for drivers. Under normal conditions, vehicle-to-user communication refers to the communication of Bluetooth and dedicated short-range communication (DSRC) \cite{bertolino2018tour}, which can also be regarded as a special case of in-vehicle communication, we define the risk of user-to-vehicle communication as
	\begin{equation}\label{key3}
		V_2 = \sum _{i=1}^{N} n_{DSRC_{i}}
	\end{equation}
	where $n_{DSRC}$ indicates the number of modules in the vehicle that communicate using the DSRC network.                                                                                                            
\end{definition} 
\begin{definition}[Vehicle-To-Vehicle Communication Risks (V2VCR)] 
	\rm Communication between two vehicles is usually done to share traffic information. 
	This communication may take place through an ad hoc network called vehicular ad hoc networks (VANETs) \cite{razzaque2013security}. 
	Then it also brings corresponding risks, such as eavesdropping on vehicle messages, forwarding errors, and even damage to the ECU \cite{demba2018vehicle}.
	We therefore define the vehicle-to-vehicle communication risk as
	\begin{equation}\label{key4}
		V_3 = \sum _{i=1}^{N} n_{VANETs_i}
	\end{equation}
	where $n_{VANETs}$ represents the number of modules that use the VANETs network for vehicle-to-vehicle communication. 
\end{definition}
\begin{definition}[Vehicle-To-Infrastructure Communication Risks (V2ICR)] 
	\rm V2I mainly refers to the sharing of information between vehicles and traffic roads to improve vehicle driving safety and improve vehicle traffic routes. 
	The following three communication technologies are mainly used VANETs, Wi-Fi and Cellular, so we define vehicle-to-infrastructure risk as 
	\begin{equation}\label{key5}
		V_4 = \sum _{i=1}^{N} \alpha_1 n_{VANETs_i} +  \alpha_2 n_{WF_i} + \alpha_3 n_{Cel_i}
	\end{equation}
	where $\alpha$ represents the degree of harm caused to the vehicle after each communication damage, $n_{VANETs}$ represents the number of modules that use the VANETs network for communication between the vehicle and the infrastructure, $n_{WF}$ represents the number of modules in the vehicle that use Wi-Fi for communication, and $n_{Cel}$ represents the number of modules in the vehicle that communicate using the cellular network.
\end{definition}

\subsubsection{Vehicle Code Complexity Risk (VCCR)}
Modern automotive systems have exceeded 100 million lines of code (LOC) and will rapidly increase to 300 million lines of code in recent years\cite{2019Autonomous}. 
However, the more lines of code, the higher the complexity, the greater the risk\cite{2011Evaluating,ChoZ2011}. 
Attackers can exploit these vehicle software vulnerabilities to compromise vehicle systems, we use vehicle code complexity risk as an indicator \cite{2011Evaluating,ChoZ2011}. 
The definitions of the metrics we mentioned as follows:

\begin{definition}[Line Of Source Code Metric Risk (LOCM)] 
	\rm $n_{LOC}$ refers to the number of source code lines of executable statements in each module of the vehicle. 
	So the complexity of source code lines is defined as
	\begin{equation}\label{key6}
		C_1 = \sum _{i=1}^{s} n_{LOC_{i}}, 
	\end{equation}
	where $s$ represents the code module in the vehicle. 
\end{definition}

\begin{definition}[Operators And Operands Complexity Metric Risk (HCM)] 
	\rm HCM is calculated based on the number of operators and operands in the vehicle software system. 
	The operator represents the operation symbol that needs to be executed in the vehicle code, and the operand represents the basic logic unit to be operated in the vehicle system code. 
	Therefore the HCM is defined as 
	\begin{equation}\label{key7}
		C_2 = \sum _{i=1}^{s} (N_1+N_2)log_2(n_1+n_2)/3000
	\end{equation}
	where $n_1$ represents the number of categories of operators, $n_2$ represents the number of categories of operands, and $N_1$ represents the number of occurrences of operators, and $N_2$ represents the number of occurrences of operands.
\end{definition}

\begin{definition}[Cyclomatic Complexity Of Code Metric Risk (CCM)] 
	\rm CCM mainly focuses on the risk measurement of the complexity of the loop structure diagram in the vehicle software module, and assigns different weights to different control structures in the vehicle software code. 
	Therefore, CCM is defined as 
	\begin{equation}\label{key8}
		C_3 = \sum _{i=1}^{s} E-H+2
	\end{equation}
	where $E$ is the number of edges in the control flow graph and $H$ is the number of nodes in the control flow graph.
\end{definition}

\subsubsection{Risk of Vehicle History Security Issues (VHIR)}

When evaluating the vulnerable components of unmotned vehicles, we also considers the risk of vehicles caused by the vehicle's previous software errors and the security attacks caused by the security attacks. 
We divides the vehicle history security subindicators into the following three categories, including the historical problems of vehicle functional security, historical issues of vehicle expected functional security, and historical problems of vehicle information security. 

\begin{definition}[Historical Problem Of Vehicle security Function(HPSF)] 
	\rm The security function problems of vehicles mainly refer to the failure of the electronic and electrical systems of the vehicle, including the failure of the cruise control system, the failure of the body control module. And the consequences of these functional security problems are often the most serious. Even need to replace parts to solve. 
	We define it as 
	\begin{equation}\label{key9}
		H_1 = \sum _{d=1}^{Y} \eta_{d} + \lambda \tau_{d}
	\end{equation}
	where $\eta_{d}$ represents the last recalled number of vehicles of this type, $\tau_{d}$ represents the last number of repairs for this vehicle type due to electrical and electronic system failures, and $\lambda$ represents the forgetting factor, which is the vehicle component that calculates how long it takes to replace the vehicle after it is purchased. 
	The longer the time, the smaller the value of $\lambda$.
\end{definition}

\begin{definition}[The Historical Problem Of Vehicle Expected Functional  Security(HPEFS)] 
	\rm Vehicle expected functional security refers to the identification of the physical environment experienced by the vehicle, which may have erroneous cognition or erroneous execution caused by unclear calibration, resulting in hidden security hazards in the vehicle, including system function limitations and environmental interference, etc. 
	So we define it as 
	\begin{equation}\label{key10}
		H_2 = \sum _{i=1}^{N} n_{spl_{i}}
	\end{equation}
	where $n_{spl}$ represents the number of accidents of vehicles due to system performance limitations.
\end{definition}

\begin{definition}[Vehicle Information Security History Problem(HPIS)] 
	\rm Vehicle information security problem refers to the vehicle system being attacked by hackers, resulting in a series of security problems in the vehicle system. 
	Hacker attacks include distributed denial of service, fuzzy attacks, and deception attacks. 
	Information security protection should be involved in the vehicle design and development stage, so we classify vehicle information security according to the core components of the vehicle and set it as
	\begin{equation}\label{key11}
		H_3 = \sum _{i=1}^{N} \varepsilon_1 n_{Ve_{i}}+ \varepsilon_2 n_{Te_{i}} + \varepsilon_3 n_{Ne_{i}} + \varepsilon_4 n_{Cl_{i}}
	\end{equation}
	where $n_{Ve}$ represents the number of attacks on the vehicle terminal, $n_{Te}$ represents the number of attacks on the vehicle terminal, $n_{Ne}$ represents the number of attacks received on the network side of the vehicle, $n_{Cl}$ represents the number of attacks on the cloud of the vehicle, and $\varepsilon$ indicates the severity of damage to the vehicle caused by the attack on various core components of the vehicle.
	
\end{definition}


\subsection{Severity Of Damage To Vehicle (D)}

In order to quantitatively evaluate the vehicle intelligent connected system, we use four indicators to comprehensively consider the damage severity of the vehicle: $D_1$ (safety, the overall impact of component damage on the vehicle), $D_2$ (privacy, the impact of vehicle data attack on the vehicle impact), $D_3$ (financial, financial loss to vehicle from damaged components), $D_4$ (operational loss, operational loss to vehicle from damaged components).
\begin{table*}[h]
	\renewcommand{\arraystretch}{1.3}
	\centering
	\caption{Severity of damage to vehicle \cite{boudguiga2015race,cui2020vera,sae2016cybersecurity}}
	\begin{tabular}{ccccc} \hline
		\textbf{Severity D}   &   \textbf{Safety $D_1$}   &   \textbf{Privecy $D_2$}  & \textbf{Financial $D_3$}  & \textbf{Operational $D_4$}    \\ \hline
		$0$ &  No injuries   &  No unauthorized access to data &  0 $\leq$ loss$\textless$100 &  No impact on performance \\ 
		$1-3$ & Light injuries    &  Access to anonymous data   &  100$\leq$loss$\textless$1000 &  Impact not detected by driver \\ 
		$4-6$ &  Severe injuries , with survival    & Identification of vehicle or driver     &  1000$\leq$loss$\textless$10000 &  Driver aware of performance degradation \\ 
		$7-9$ &  Life threatening, possible death     &  Driver or vehicle tracking   &  loss$\geq$10000 &  Significant impact on performance \\ \hline
		
	\end{tabular}
	
	\label{table:severity}
\end{table*} 
We list Table \ref{table:severity} according to the definition of vehicle damage severity in SAE J3061 \cite{sae2016cybersecurity}, RACE\cite{boudguiga2015race} and VeRA\cite{cui2020vera}. 
In this paper, we use the maximum value of the above severity vector to represent the vehicle damage severity level. 
For example, assuming the above severity vector [6, 3, 5, 4] (which is $D_1=6, D_2=3, D_3=5, D_4=4$), it can be known that the severity level of vehicle damage is $D=6$.

However, we cannot accurately predict the absolute degree of damage for each influencing factor. 
Therefore, based on the safety data and financial data disclosed by major vehicle data websites (please see for sources 
\href{http://n3.datasn.io/data/api/v1/n3_chennan/peijian/main/list/?app=html-bunker, https://www.crysys.hu/research/vehicle-security/}{n3.datasn.io}, \href{https://www.crysys.hu/research/vehicle-security/, https://www.crysys.hu/research/vehicle-security/}{www.crysys.hu} and some previous papers \cite{2008Vehicle, cui2020vera,boudguiga2015race}).
We constructed a matrix to reflect the relative importance of the damage degree of each impact vehicle index, as shown in Table \ref{table:VSR}, \ref{table:ECR}, \ref{table:VCR}, \ref{table:VCCR}, \ref{table:VHIR}, and adopted the following core algorithm model to dynamically adjust the severity of vehicle damage through each influencing factor, making our vehicle intelligent connected system evaluation model more accurate.

\begin{table}[h]
	\renewcommand{\arraystretch}{1.2}
	\centering	
	\caption{VSR first-order factor damage severity matrix}
	\begin{tabular}{c|cccc} \hline
		\textbf{VSR}   &   \textbf{ECR}    &   \textbf{VCR}  & \textbf{VCCR} & \textbf{VHIR}  \\ \hline
		\textbf{ECR} &  1.000   &  1.143  & 2.000 &  2.667 \\ 
		\textbf{VCR}&  0.875    &   1.000   &  1.750 &  2.333 \\ 
		\textbf{VCCR}&  0.500     &  0.571    &  1.000 &  1.333 \\ 
		\textbf{VHIR} &  0.375     &  0.429   &  0.750 &  1.000 \\ \hline
		
	\end{tabular}
	
	\label{table:VSR}
\end{table} 

\begin{table}[h]
	\renewcommand{\arraystretch}{1.2}
	\centering
	\caption{ECR indicator damage degree severity matrix}
	\setlength{\tabcolsep}{0.6mm}{
		\begin{tabular}{c|cccccccc} \hline
			\textbf{ECR}   &   \textbf{EMCR}   &   \textbf{TCCR}  & \textbf{EPCR}  & \textbf{ESCR} & \textbf{MRCR} & \textbf{ACCR} & \textbf{LKCR} & \textbf{AECR}\\ \hline
			\textbf{EMCR} &  1.000   &  1.167  &  0.778 &  0.875  &  1.400  &  1.750  &  1.400  &  0.778 \\ 
			\textbf{TCCR} &  0.857    &  1.000     &  0.667 &  0.750   &  1.200  &  1.500  &  1.200  &  0.667 \\
			\textbf{EPCR} &  1.286     &  1.500    &  1.000 &  1.125   &  1.800  &  2.250  &  1.800  &  1.000\\ 
			\textbf{ESCR} &  1.143     &  1.333   &  0.889 &  1.000   &  1.600  &  2.000  &  1.600 &  0.889\\ 
			\textbf{MRCR} &  0.714     &  0.833   &  0.556 &  0.625   &  1.000  &  1.250  &  1.000  &  0.556\\ 
			\textbf{ACCR} &  0.571     &  0.667   &  0.444 &  0.500   &  0.800  &  1.000 &  0.800  &  0.444\\ 
			\textbf{LKCR} &   0.714     &  0.833   &  0.556 &  0.625   &  1.000  &  1.250  &  1.000  &  0.556\\ 
			\textbf{AECR} & 1.286     &  1.500    &  1.000 &  1.125   &  1.800  &  2.250  &  1.800  &  1.000\\ \hline
			
		\end{tabular}
	}
	\label{table:ECR}
\end{table} 

\begin{table}
	\renewcommand{\arraystretch}{1.2}
	\centering
	\caption{VCR indicator damage degree severity matrix}
	\begin{tabular}{c|cccc} \hline
		\textbf{VCR}   &   \textbf{IVCR}    &   \textbf{U2VCR}  & \textbf{V2VCR}  & \textbf{V2ICR}    \\ \hline
		\textbf{IVCR} &  1.000  &  1.600  &  2.667 &  1.143\\ 
		\textbf{U2VCR} &  0.625    &  1.000     &  1.667 &  0.714 \\ 
		\textbf{V2VCR} &  0.375     &  0.600    &  1.000 &  0.429 \\ 
		\textbf{V2ICR} &  0.875     &  1.400   &  2.333 &  1.000 \\ \hline
		
	\end{tabular}
	
	\label{table:VCR}
\end{table} 

\begin{table}
	\renewcommand{\arraystretch}{1.2}
	\centering	
	\caption{VCCR indicator damage degree severity matrix}
	\begin{tabular}{c|ccc} \hline
		\textbf{VCCR}   &   \textbf{LOCM}    &   \textbf{HCM}  & \textbf{CCM}     \\ \hline
		\textbf{LOCM} &  1.000    &  0.333     &  0.200 \\ 
		\textbf{HCM} &  3.000    &  1.000   &  0.600 \\ 
		\textbf{CCM} &  5.000     &  1.667   &  1.000 \\ \hline
		
	\end{tabular}
	
	\label{table:VCCR}
\end{table} 

\begin{table}[h]
	\renewcommand{\arraystretch}{1.2}
	\centering
	\caption{VHIR indicator damage degree severity matrix}
	\begin{tabular}{c|ccc} \hline
		\textbf{VHIR}   &   \textbf{HPSF}  &   \textbf{HPESF}  & \textbf{HPIS}     \\ \hline
		\textbf{HPSF} &  1.000    &  2.667     &  1.600 \\ 
		\textbf{HPESF} &  0.375     &  1.000    &  0.600 \\ 
		\textbf{HPIS} &  0.625    &  1.667   &  1.000 \\ \hline
		
	\end{tabular}
	
	\label{table:VHIR}
\end{table} 

\section{Establishment Of An Evaluation Model}
\label{sec:algorithm}
This paper adopts the multi-index system to evaluate the risk of the vehicle software integrated system, which mainly includes evaluation structure, risk model establishment and calibration evaluation. 
The evaluation structure has been explained in the third section, and the establishment of the risk model in the fourth section will combine the 15 sub-indicators selected from the above four second-order indicators of vehicle software safety with the vehicle risk data map.
Beside it, the calibration evaluation is based on the use of I-FAHP method performs the calibration of the subjective weights of the evaluation factors.
In the following, we will introduce the vehicle risk assessment model we established in detail.


\subsection{Evaluation Of Architecture}


In the ECU coupling of the vehicle, the relevant risk index includes the ECU coupling in the major systems (such as EMCR, TCCR, EPCR, ESCR). 
For vehicle communication risks, it includes IVCR, U2VCR, V2VCR, V2ICR. 
The code complexity risk represents the LOCM, HCM, CCM of the code in the vehicle intelligent connected system. 
Finally, the safety history risk of the vehicle is expressed as HPSF, HPEFS, HPIS. Therefore, the vehicle intelligent connected system risk can be conceptually described using equation (12).
\begin{equation}
	\begin{split}
		VSR= ECR \otimes VCR \otimes VCCR \otimes VHIR 
	\end{split}
\end{equation}
where the symbol $\otimes$ is only used to represent the overlay analysis of the vehicle software system risk equation. 
We consider that different influencing factors have different degrees of influence on the risk of vehicle intelligent connected system, so the combination of different evaluation factors and their corresponding weights is used as the risk assessment of vehicle intelligent connected system, and VSR is redefined using equation (13).
\begin{equation}
	\begin{split}
		VSR= \omega_E(\sum _{i=1}^{n}e_iE_i)_{ECR} \otimes \omega_V(\sum _{j=1}^{n}v_jV_j)_{VCR} \\
		\otimes \omega_C(\sum _{k=1}^{n}c_kC_k)_{VCCR} \otimes \omega_H(\sum _{l=1}^{n}h_lH_l)_{VHIR} \\
	\end{split}
\end{equation}
Among them, $\omega_E$, $\omega_V$, $\omega_C$ and $\omega_H$ are the vehicle ECU coupling risk index, vehicle communication risk index, vehicle code complexity risk index and vehicle history information security, respectively event risk weight for the problem index.

According to the above evaluation structure, we analyzed and dealt with each impact sub-indicator. 
In order to obtain accurate evaluation results, we normalize the above-mentioned vehicle influencing factors before superimposing each influencing indicator, and the value range of the normalized vehicle software safety influencing indicators is $\left[ 0, 1 \right]$. The normalization method is shown in equation (14).
\begin{equation}
	\begin{split}
		\beta_{ij}= \frac{\alpha _{max} - \alpha _{ij}}{\alpha _{max} - \alpha _{min}}
	\end{split}
\end{equation}
where $ \beta _{ij} $ is the standardized value of vehicle influencing factors; $\alpha _{max}$ and $\alpha _{min}$ correspond to the maximum and minimum values of each influencing sub-factor of the vehicle system. $\alpha _{ij}$ represents the original value of the second-order influence element in the vehicle.

To assess the security risk of vehicle software, we introduce the Interval-Fuzzy Analytic Hierarchy Process (I-FAHP) into the above assessment framework. 
The I-FAHP method uses interval fuzzy numbers on the basis of AHP to represent the importance of the second-order influencing factors of each vehicle in the judgment matrix.
In the next subsection, we describe in detail the I-FAHP method used.

\subsection{Describes The I-FAHP Method} 

The interval FAHP weight calculation is based on the above safety risk assessment structure of vehicle intelligent connected system. 
The judgment matrix ($\varphi$) of each evaluation factor can be expressed by equation (15) :

\quad{\bfseries $\varphi = [\varphi^{-} , \varphi^{+}]$} = $(\varphi_{ij})_{m \times m}$ = $(\varphi_{ij} ^{-} , \varphi_{ij} ^{+})_{m \times m}$\\
\begin{equation}
	=\begin{cases}
		\;\;[1, 1] \quad [\varphi_{12} ^{-} , \varphi_{12} ^{+}] \cdots [\varphi_{1i} ^{-} , \varphi_{1i} ^{+}] \cdots [\varphi_{1m} ^{-} , \varphi_{1m} ^{+}]  \\
		[\frac{1}{\varphi_{12} ^{-} }, \frac{1}{\varphi_{12} ^{+} }] \quad  
  [1, 1] \;\;\;\cdots [\varphi_{2i} ^{-} , \varphi_{2i} ^{+}] \cdots [\varphi_{2m} ^{-} , \varphi_{2m} ^{+}]  \\
		
		\;\quad \cdots \quad \quad \;\cdots \quad \;\;  \cdots \quad  \;\cdots \quad  \cdots \quad \ \, \cdots \\
		[\frac{1}{\varphi_{1i} ^{-} }, \frac{1}{\varphi_{1i} ^{+} }]\;\; [\frac{1}{\varphi_{2i} ^{-} }, \frac{1}{\varphi_{2i} ^{+} }]\;\, \cdots \;\; [1, 1]  \;\; \cdots [\varphi_{im} ^{-} , \varphi_{im} ^{+}]  \\
		
	\;\quad \cdots \quad \quad \;\cdots \quad \;\;  \cdots \quad  \;\cdots \quad  \cdots \quad \ \, \cdots \\
		
		[\frac{1}{\varphi_{1m} ^{-} }, \frac{1}{\varphi_{1m} ^{+} }] [\frac{1}{\varphi_{2m} ^{-} }, \frac{1}{\varphi_{2m} ^{+} }] \cdots [\frac{1}{\varphi_{im} ^{-} }, \frac{1}{\varphi_{im} ^{+} }] \cdots \;\;\,[1, 1]  \\
\end{cases} \end{equation}

The interval judgment matrix ($\varphi$) is divided into two parts: the lower limit judgment matrix ($\varphi^{-}$) and the upper limit judgment matrix ($\varphi^{+}$). $\varphi_{ij} ^{-}$ and $\varphi_{ij} ^{+}$ are elements in the judgment matrix $\varphi$, representing the relative importance of factor $\varphi_{i}$ to factor $\varphi_{j}$. 
When $\varphi_{i}$ is much more important than $\varphi_{j} $, $\varphi_{ij}$ is set to 9, and $\varphi_{ji}$ is set to 1/9, where $ 1/9 \leq \varphi_{ij} ^{-} \leq  \varphi_{ij} ^{+} \leq 9$. 
The weight vector of interval judgment matrix can be calculated according to equation (16).

\begin{equation}
	\begin{cases}
		
		$$
		\vec {\gamma} =[\gamma_{1} , \gamma_{2}] = [\mu \gamma ^{-} , \nu\gamma^{+} ]
		$$  \\
		\mu =  \sqrt{\sum _{j=1}^{m}{\frac{1}{\sum _{i=1}^{m}\varphi_{ij}^{+}  }}}     \\
		\nu =  \sqrt{\sum _{j=1}^{m}{\frac{1}{\sum _{i=1}^{m}\varphi_{ij}^{-}  }}}
	\end{cases} 
\end{equation}

$\gamma ^{-}$ and $\gamma ^{+}$ are the weight of the lower bound matrix and the weight of the upper bound matrix, respectively. $\gamma_{1}$ and $\gamma_{2}$ are the weight of the interval judgment matrix, when the weight of the interval FAHP method repeatedly covers the weight of the original AHP method, the interval judgment matrix $\vec {\gamma}$ is considered reasonable, where $\gamma$ is the weight calibrated by the AHP method.
As shown in equation (17).
\begin{equation}
	\gamma_{1} \leq \gamma  \leq \gamma_{2}
\end{equation}

\subsection{Describes The FCA Method} 

The FCA method is mainly used to modify the weights obtained from the I-FAHP method and to analyze the relative relationship between the vehicle data samples and the above-mentioned vehicle evaluation criteria.
Assuming that there are $x$ vehicles with evaluation sample data and $y$ impact indicators, the eigenvectors formed are:
\begin{equation}
	{ N = (n_{ij})_{x \times y}}=\begin{bmatrix}
		n_{11} & n_{12} & \cdots &n_{1y} \\
		n_{21} & n_{22} & \cdots &n_{2y}\\
		\vdots & \vdots & \ddots &\vdots\\
		n_{x1} & n_{x2} & \cdots &n_{xy}
		
	\end{bmatrix}
\end{equation}

The $n_{ij}$ in the matrix elements is the eigenvalue of the vehicle influence factor $j$ to the vehicle data sample $i$. 
In order to eliminate the influence of different data sizes in the vehicle, we normalize each element in the matrix $N$ according to equation (14).

Assuming that there are $y$ second-order influencing factors of $x$ evaluation sample data in the vehicle, $a$ class can be used for clustering, so the formed fuzzy clustering matrix $P$ can be represented by equation (19):

\begin{equation}
	{ P = (p_{ei})_{a \times x} } = \begin{bmatrix}
		p_{11} & p_{12} & \cdots &p_{1x} \\
		p_{21} & p_{22} & \cdots &p_{2x}\\
		\vdots & \vdots & \ddots &\vdots\\
		p_{a1} & p_{a2} & \cdots &p_{ax}
	\end{bmatrix}
\end{equation}
where $p_{ei}$ represents the relative degree of correlation between vehicle data sample $i$ and category $a$, $0\leq p_{ei} \leq  1$ , $\sum _{e=1}^{a}p_{ei} = 1$. $x$ represents the number of vehicle sample data, and $e$ is the number of categories. Among them, the eigenvalues of a class $y$ index are expressed as cluster centers, so the fuzzy class center matrix S can be expressed by equation (20):
\begin{equation}
	{S = (s_{ja})_{y \times e} } = \begin{bmatrix}
		s_{11} & s_{12} & \cdots &s_{1e} \\
		s_{21} & s_{22} & \cdots &s_{2e}\\
		\vdots & \vdots & \ddots &\vdots\\
		s_{e1} & s_{e2} & \cdots &s_{ye}
	\end{bmatrix}
\end{equation}
where $s_{ja}$ represents the relative degree of correlation between the vehicle index $j$ and class $a$,  $0\leq s_{ja} \leq  1$.

Based on the above matrices $P$, 
and $S$,the calculation equation of the eigenvalue class of the vehicle evaluation sample is as follows equation(21):

\begin{equation}
	R_{i} = \sum _{e=1}^{a}p_{ei}\cdot e  
\end{equation}
where $R_{i}$ is the eigenvalue category in the vehicle data sample $i$. 
According to the value of $R_{i}$, the risk class of the vehicle data evaluation sample can be determined.

\subsection{Comprehensive Evaluation Model} 

We emphasize again that the main contribution of this paper is to use the FCA and FAHP methods to assess the risk of the vehicle intelligent connected system before the vehicle is on the ground. 
There is no better assessment method in the field of vehicle software risk assessment. 
In the above FCA, we can get the corresponding sensitivity coefficient $\rho$, where $\rho$ is mainly used to modify the weight obtained by the I-FAHP method, and the modified weight $\varphi$ is shown in equation (22):

\begin{equation}
	\varphi = \rho \varphi_{s} + (1-\rho) \varphi_{o} 
\end{equation}
where $0\leq \rho \leq  1$, $\varphi_{s}$ represents the subjective weight, which can be calibrated by the I-FAHP method; $\varphi_{o}$ represents the objective weight, which is used to reflect the characteristics of the vehicle evaluation data sample. 

Then we use the projection pursuit (PP) method to calibrate the value of $\varphi_{oj}$. 
Assuming that $n_{ij}$ is the coefficient $j$ of the vehicle evaluation data sample $N$ index $i$, the best calibration vector is $\vec {l}$, and equation can be obtained according to the PP, as follows equation (23):
\begin{equation}
	\phi_j = \sum _{j=1}^{y}\vec {l}_jz_{ij} ,\quad i \in [1, x], j \in [1, y]
\end{equation}
where x represents the number of evaluation sample data, and y represents the number of impact indicators.
The direction vector $\vec {l}_j$ represents the characteristics of the vehicle data sample, $\vec {l}_j = (l_1, l_2, l_3 \cdots, l_y)$, and $z_{ij}$ represents the normalization factor $j$ of the vehicle influencing factor $i$. 
In order to obtain the optimized $\vec {l}_j$, the objective function $f(l)$ is established as the following equation (24):

\begin{equation}
	\begin{cases}
		
		$$
		A_{\phi} =(\frac{\sum _{i=1}^{x}(\phi_{i}-\bar \phi)}{x-1})
		$$  \\
		
		$$
		b_{ia} = \rvert \rvert  \phi_{i} - \phi_{a} \rvert \rvert
		$$  \\

		$$
		B_{\phi} =\sum _{i=1}^{x}\sum _{a=1}^{x}(b-b_{ia})\cdot f(b-b_{ia})
		$$  \\
		
		maxR(l) = A(l) \cdot B(l)
	\end{cases} 
\end{equation}
In the above equation, $\bar \phi$ represents the average value of $\phi_{i}$, and $A_{\phi}$ represents the standard deviation among the vehicle data evaluation samples, $B_{\phi}$ represents the local density function, and $f(b-b_{ia})$ is a unit step function, as follows equation (25):

\begin{equation}
	f(b-b_{ia}) =  \begin{cases}
		1,& b-b_{ia} \geq 0 \\
		0,& b-b_{ia} \textless 0 
		
	\end{cases} 
\end{equation}

\section{Experimental results} \label{sec:experiments}

In this section, we conduct experimental case studies on OpenPilot \cite{fontana2021self}.
OpenPilot is a relatively mature $L_2$ assistance-oriented open source autonomous driving system \cite{fontana2021self}, which implements assisted driving functions such as adaptive cruise, lane keeping assistance, driver status monitoring based on an end-to-end model. 
In addition, OpenPilot only needs a mobile phone and Qualcomm Snapdragon 821 chip, equipped with OpenPilot software, to achieve L2+ autonomous driving. 
Last but not least, in the evaluation report based on consumer usage evaluation in 2020, \href{https://data.consumerreports.org/wp-content/uploads/2020/11/consumer-reports-active-driving-assistance-systems-november-16-2020.pdf}{the comprehensive score of OpenPilot is as high as 78, ranking first.} Therefore, we comprehensively consider and use OpenPilot to conduct experiments. 
We will describe the test steps of our proposed evaluation method on OpenPilot\cite{fontana2021self}, including the calibration of the weights of each vehicle influencing factor in I-FAHP in section 5.1, the analysis of vehicle evaluation indicators in section 5.2, and finally the description of the evaluation results of vehicle indicators on OpenPilot in section 5.3.
Our experiments were performed on a computer running Windows 10 enterprise with a 64-bit Core(TM) i5-9400, a 2.90GHz Intel Core processor, and 16GB of main memory.

\subsection{Evaluate Model Weight Calibration}

The weight calibration of the vehicle software integrated system risk assessment model is mainly divided into two parts. 
The first part is to calibrate the weights by the I-FAHP method, and the second part is to use the FCA method to modify the weights obtained by I-FAHP. 
In this paper, the data source of each evaluation factor is as follows.
(1) $ E_{i} $ represent respectively: $ E_{1} $ stands for EMCR, $ E_{2} $ stands for TCCR, $ E_{3} $ stands for EPCR, $ E_{4} $ stands for ESCR, $ E_{5} $ stands for MRCR, $ E_{6} $ stands for ACCR, $ E_{7} $ stands for LKCR, $ E_{8} $ stands for AECR, the data source of $ E_{1} $ to $ E_{5} $ is Major vehicle intelligent connected system platforms and the data source of $ E_{6} $ to $ E_{8} $ is  Active Driving Assistance Systems: Test Results and Design Recommendations.
(2) $ V_{i} $ represent respectively: $ V_{1} $ stands for IVCR, $ V_{2}$ stands for U2VCR, $ V_{3} $ stands for V2VCR, $ V_{4} $ stands for V2ICR, the data source of $ V_{i} $ is Communication Architecture of Major Vehicles. 
(3) $ C_{i} $ represent respectively: $ C_{1} $ stands for LOCM, $ C_{2}$ stands for HCM, $ C_{3} $ stands for CCM, the data source of $ C_{i} $ is Core source code of major vehicle systems.
(4) $ H_{i} $ represent respectively: $ H_{1} $ stands for HPSF and the data source of $ H_{1} $ is Maintenance records of major vehicles, $ H_{2}$ stands for HPESF and the data source of $ H_{2} $ is Traffic safety accidents of major vehicles, $ H_{3} $ stands for HPIS and the data source of $ H_{3} $ is Attack data of major vehicle systems.

\subsubsection{Weights Obtained By I-FAHP}

According to the evaluation structure shown in Fig. \ref{fig:structure1} above, we use the I-FAHP method to calibrate the weights of the vehicle evaluation impact indicators. 
In order to obtain the interval fuzzy weight, the original AHP weight should be calibrated first, which uses the form of pairwise comparison of vehicle influence factors to express the relative importance of vehicle influence factors. 
Table \ref{table:VSR}, \ref{table:ECR}, \ref{table:VCR}, \ref{table:VCCR}, \ref{table:VHIR} lists the data of our major vehicle impact indicators, and then based on the consistency judgment matrix, the corresponding interval judgment matrix can be obtained. 
Table \ref{table:7} lists the weights of the vehicle impact indicators calibrated by the I-FAHP method, where the AHP weights are within the I-FAHP weights, which confirms the applicability of our I-FAHP judgment matrix.

\begin{table*}
	\centering
	\setlength{\tabcolsep}{10pt}
	\renewcommand{\arraystretch}{1.3}
	\caption{Weights of vehicle intelligent connected system impact indicators calibrated according to I-FAHP and FAHP-FCA}
	\setlength{\tabcolsep}{2.5mm}{
		\begin{tabular}{cccccccc}
			\hline
			\multicolumn{3}{l}{Index layer} &Sub-index layer&\multicolumn{4}{l}{} \\ \hline
			Index & AHP&I-FAHP&Factor&AHP&I-FAHP&FCA-AHP&FCA-I-FAHP \\[0.15cm]\hline
			\multirow{8}{*}{ECR} &\multirow{8}{*}{0.3636} &\multirow{8}{*}{[0.2949,0.3702]} &EMCR&0.1321&[0.0861,0.1810]&0.130432&[0.087192,0.176398]\\
			& & &TCCR&0.1132&[0.0785,0.1333]&0.113956&[0.081338,0.132850] \\
			&& & EPCR&0.1698&[0.1351,0.2043]&0.170304&[0.137686,0.202734]\\
			& & &ESCR&0.1510&[0.1180,0.1681]&0.150778&[0.119758,0.166852]\\
			& & &MRCR&0.0943&[0.0702,0.1048]&0.096154&[0.073500,0.106024] \\
			& & &ACCR&0.0755&[0.0495,0.0757]&0.078962&[0.054522,0.079150]\\
			& & &LKCR&0.0943&[0.0824,0.0993]&0.091834 &[0.080648,0.096534]\\
			& & &AECR&0.1698&[0.1380,0.2719]&0.167580 &[0.137688,0.263554]\\ \hline
			\multirow{4}{*}{VCR}&\multirow{4}{*}{0.3182}&\multirow{4}{*}{[0.2778,0.4013]}&IVCR&0.3478&[0.2935,0.3767]&0.352864&[0.301822,0.380030] \\
			& & &U2VCR&0.2174&[0.1938,0.2522]&0.214094&[0.191910,0.246806] \\
			& & &V2VCR&0.1305&[0.1241,0.1405]&0.127728 &[0.121712,0.137128] \\
			& & &V2ICR&0.3043&[0.2643,0.3548]&0.305314&[0.267714,0.352784]\\ 
			\hline
			\multirow{3}{*}{VCCR}	 &\multirow{3}{*}{0.1818}&\multirow{3}{*}{[0.1527,0.1911]}&LOCM&0.1111&[0.1089,0.1136]&0.117526 &[0.115458,0.119876]\\
			& & &HCM&0.3333&[0.3097,0.3713]&0.333060 &[0.310876,0.368780]\\
			& & &CCM&0.5556&[0.5094,0.5834]&0.549414&[0.505986,0.575546]\\ 
			\hline
			\multirow{3}{*}{VHIR}	&\multirow{3}{*}{0.1364}&\multirow{3}{*}{[0.1211,0.1435]}&HPSF&0.5000&[0.4524,0.5258]&0.497198&[0.452454,0.521450]\\
			& & &HPESF&0.1875&[0.1786,0.2060]&0.183894 &[0.175528,0.201284]\\
			& & &HPIS&0.3125&[0.2814,0.3541]&0.318908&[0.289674,0.358012]\\  \hline
	\end{tabular}}
	\label{table:7}
\end{table*} 

\subsubsection{FCA Calibration Interval Weights}

According to the subjective weight of vehicle influencing factors calibrated by I-FAHP method, the sensitivity coefficient $\rho$ calculated by FCA method is used to correct the subjective weight. Then, through the obtained vehicle sensitivity coefficient and the target weight of the vehicle impact index, the corrected weight of each influencing factor of the vehicle can be obtained by using the above eqution (22), and the corrected weight is listed in the second right column of the Table \ref{table:7} . 
The weight of ECR is $E_i=(E_1, E_2, E_3, E_4, E_5, E_6, E_7, E_8)$ =
([0.087192,0.176398],[0.051338,0.13285],[0.137686,0.202734], [0.119758,0.166852],[0.07350,0.106024],[0.054522,0.079150], [0.080648,0.096534],[0.137688,0.263554]), and the weight of VCR is $V_j=(V_1,V_2,V_3,V_4)$ =
([0.301822,0.38003], [0.19191,0.246806],[0.121712,0.137128],[0.267714,0.352784]); the weight of VCCR is $C_k=(C_1, C_2, C_3)$= 
([0.115458, 0.119876],[0.310876,0.36878],[0.505986,0.575546]) and the weight of VHIR is $ H_l=(H_1, H_2, H_3)$ = ([0.452454,0.52145], [0.175528,0.201284],[0.289674,0.358012]).



\subsection{Analysis of Influencing Factors of Vehicles}

Then we normalized the vehicle impact indicators we defined above based on the risk assessment forms in ISO 26262\cite{PalWH2011} and ISO/SAE 21434\cite{MacSV2020}, and defined a brand new vehicle software integrated system risk assessment matrix.
Due to the need to increase the number of levels in HARA and TARA from 4 to 5, we chose to build a new matrix entirely instead of first normalizing the tables in HARA and TARA, then increasing the number of levels and adjusting again.
The newly proposed rating mapping table for each impact indicator is shown in Table \ref{table:8}, \ref{table:9}, \ref{table:10}, \ref{table:11}. 
Then, according to the above tables, using the overlapping interval lower limit method combined with the corresponding weights listed in the table \ref{table:7}, the rating mapping of the first-order impact indicators in the table \ref{table:12} can be obtained.
Among them, $ECR_{nor}$, $VCR_{nor}$, $VCCR_{nor}$, and $VHIR_{nor}$ represent the sum of normalized parameters of ECR, VCR, VCCR, and VHIR respectively.
Finally, according to the rating mapping table and table \ref{table:12} of the above vehicle impact indicators, combined with the objective function equation (13), the rating mapping of the vehicle intelligent connected system risk assessment matrix is constructed (as shown in Table \ref{table:13}).


\begin{table}[h]
	\renewcommand{\arraystretch}{1.2}
	\centering
	\caption{ECR Influencing Factors Evaluation Criteria}
	\setlength{\tabcolsep}{0.8mm}{
		\begin{tabular}{cccccc} \hline
			\textbf{ECR}   &  \textbf{Normal}    &   \textbf{Slight} & \textbf{Slightly Serious} & \textbf{Serious} & \textbf{Extremely Serious}  \\ \hline
			$E_1$ &  0-0.30  &   0.30-0.45  &   0.45-0.60 &   0.60-0.80 &   0.80-1.00\\ 
			$E_2$ &  0-0.35  &  0.35-0.55  &  0.55-0.70 &  0.70-0.85&   0.85-1.00\\ 
			$E_3$ &  0-0.20    &  0.20-0.35     &  0.35-0.55 &  0.55-0.70 &   0.70-1.00\\ 
			$E_4$ &  0-0.25     &  0.25-0.40    &  0.40-0.55 &  0.55-0.75&   0.75-1.00 \\
			$E_5$ &  0-0.40     &  0.40-0.55    &  0.55-0.80 &  0.80-0.90&   0.90-1.00\\
			$E_6$ &  0-0.45     &  0.45-0.65   &  0.65-0.85 &  0.85-0.92&   0.92-1.00 \\
			$E_7$ &  0-0.40     &  0.40-0.55    &  0.55-0.80 &  0.80-0.90&   0.90-1.00 \\
			$E_8$ &  0-0.20     &  0.20-0.35    &  0.35-0.50 &  0.50-0.70&   0.70-1.00 \\
			\hline
			
		\end{tabular}
	}
	\label{table:8}
\end{table}
\begin{table}[h]
	\renewcommand{\arraystretch}{1.2}
	\centering
	\caption{VCR Influencing Factors Evaluation Criteria}
	\setlength{\tabcolsep}{0.8mm}{
		\begin{tabular}{cccccc} \hline
			\textbf{VCR}   &  \textbf{Normal}    &   \textbf{Slight} & \textbf{Slightly Serious} & \textbf{Serious} & \textbf{Extremely Serious}  \\ \hline
			$V_1$ &  0-0.30  &   0.30-0.40  &   0.40-0.60 &   0.60-0.80 &   0.80-1.00\\ 
			$V_2$ &  0-0.35  &  0.35-0.55  &  0.55-0.75 &  0.75-0.90&   0.90-1.00\\ 
			$V_3$ &  0-0.45    &  0.45-0.60     &  0.60-0.75 &  0.75-0.85 &   0.85-1.00\\ 
			$V_4$ &  0-0.35     &  0.35-0.45    &  0.45-0.70 &  0.70-0.90&   0.90-1.00\\  \hline
			
		\end{tabular}
	}
	\label{table:9}
\end{table} 

\begin{table}[h]
	\renewcommand{\arraystretch}{1.2}
	\setlength{\tabcolsep}{5.8pt}
	\centering
	\caption{VCCR Influencing Factors Evaluation Criteria}
	\setlength{\tabcolsep}{0.8mm}{
		\begin{tabular}{cccccc} \hline
			\textbf{VCCR}   &  \textbf{Normal}    &   \textbf{Slight} & \textbf{Slightly Serious} & \textbf{Serious} & \textbf{Extremely Serious}  \\ \hline
			$C_1$ &  0-0.40  &   0.40-0.60  &   0.60-0.80 &   0.80-0.90 &   0.90-1.00\\ 
			$C_2$ &  0-0.20 &  0.20-0.45  &  0.45-0.70 &  0.70-0.85&   0.85-1.00\\ 
			$C_3$ &  0-0.15    &  0.15-0.40     &  0.40-0.65 &  0.65-0.80 &   0.80-1.00\\ 
			\hline
			
		\end{tabular}
	}
	\label{table:10}
\end{table} 

\begin{table}[h]
	\renewcommand{\arraystretch}{1.2}
	\setlength{\tabcolsep}{4pt}
	\centering
	\caption{VHIR Influencing Factors Evaluation Criteria}
	\setlength{\tabcolsep}{0.8mm}{
		\begin{tabular}{cccccc} \hline
			\textbf{VHIR}   &  \textbf{Normal}    &   \textbf{Slight} & \textbf{Slightly Serious} & \textbf{Serious} & \textbf{Extremely Serious}  \\ \hline
			$H_1$ &  0-0.20  &   0.20-0.50  &   0.50-0.70 &   0.70-0.90 &   0.90-1.00\\ 
			$H_2$ &  0-0.40  &  0.40-0.55  &  0.55-0.75 &  0.75-0.85&   0.85-1.00\\ 
			$H_3$ &  0-0.30  &  0.30-0.45     &  0.45-0.70 &  0.70-0.80 &   0.80-1.00\\ 
			\hline		
		\end{tabular}
	}
	\label{table:11}
\end{table} 

\begin{table*}[h]
	\renewcommand{\arraystretch}{1.2}
	\setlength{\tabcolsep}{8pt}
	\centering
	\caption{Evaluation criteria for the first-order influencing factors of vehicles}
	\begin{tabular}{cccccc} \hline
		\textbf{Rask Rating}   &  \textbf{Normal}    &   \textbf{Slight} & \textbf{Slightly Serious}  & \textbf{Serious} & \textbf{Extremely Serious}   \\ \hline
		ECR$_{nor}$ &  0-0.2258343  &   0.2258343-0.3484771  &   0.3484771-0.4962365 &   0.4962365-0.6103646&   0.6103646-1.0\\ 
		VCR$_{nor}$ &  0-0.3061854  &  0.3061854-0.4197778  &  0.4197778-0.6037095 &  0.6037095-0.7585744&   0.7585744-1.0\\ 
		VCCR$_{nor}$ &  0-0.1842563    &  0.1842563-0.4115634     &  0.4115634-0.6388705 &  0.6388705-0.7729456 &   0.7729456-1.0\\ 
		VHIR$_{nor}$ &  0-0.2476042     &  0.2476042-0.4531207    &  0.4531207-0.6511356 &  0.6511356-0.7881466&   0.7881466-1.0 \\  \hline
		
	\end{tabular}
	
	\label{table:12}
\end{table*} 
Then, on the basis of the risk matrix, we innovatively defined three states of the vehicle intelligent connected system, steady state, critical state, and dangerous state, and further proposed corresponding safety measures.

In Table \ref{table:13}, we have divided the risk rating of the vehicle intelligent connected system. 
It can be seen that when the VSR is greater than or equal to 0.758, the vehicle intelligent connected system is at serious risk. 
At this time, the vehicle state is in a dangerous state. 
The integrated system should be repaired; when the VSR is greater than or equal to 0.506 or less than 0.758, the vehicle intelligent connected system is at a slight serious risk, and the vehicle state is critical at this time. 
At this time, specific analysis should be carried out according to the vulnerable components in the first stage, such as the ECU of the first stage influencing factors if the coupling risk is at a serious level, the vehicle ECU system should be rectified until the ECU coupling risk is at normal or slight level and VSR is less than 0.506, and the vehicle intelligent connected system is at normal or slight risk level; and when VSR is greater than or equal to 0 or less than 0.506, the vehicle the intelligent connected system is in normal or slight risk. 
At this time, the vehicle state is in a steady state. 
At this time, it has reached the quality management level of the vehicle intelligent connected system \cite{hommes2012review}, and there is no need to rectify the vehicle intelligent connected system.

\begin{table*}
	\tiny
	\centering
	\caption{vehicle intelligent connected system risk assessment matrix}
	\setlength{\tabcolsep}{0.01mm}{
		\resizebox{\textwidth}{!}{
			\renewcommand{\arraystretch}{1.7}
			\begin{tabular}{|c|c|ccccc|ccccc|ccc|cc}
				\hline
				& VCR &\multicolumn{5}{c|}{0-0.306} & \multicolumn{5}{c|}{0.306-0.42}&\multicolumn{3}{c|}{0.42-0.604 } & &  \\ \hline
				
				ECR &\diagbox{\makebox[0.035\textwidth][c]{VHIR}}{\makebox[0.035\textwidth][c]{VCCR}}  & 0-0.184  &  0.184-0.412& 0.412-0.639  &  0.639-0.773&0.773-1&0-0.184& 0.184-0.412 & 0.412-0.639 &  0.639-0.773&0.773-1& 0-0.184 &0.184-0.412  & 0.412-0.639 & & \\ \hline
				
				\multirow{5}{*}{ 0-0.226} &  $0-0.248$  &[0,0.277] & [0.277,0.307] & [0.307,0.335] & [0.335,0.355] & [0.335,0.385] & [0.085,0.323] & [0.323,0.352] & [0.352,0.381] & [0.381,0.4] & [0.4,0.431] & [0.117,0.397] & [0.397,0.426] & [0.426,0.455] & & \\
				$ $ &  $0.248-0.453$  &[0.028,0.277] & [0.277,0.35] & [0.35,0.379] & [0.379,0.398] & [0.398,0.429] & [0.113,0.366] & [0.366,0.396] & [0.396,0.424] & [0.424,0.444] & [0.444,0.474] & [0.145,0.44] & [0.44,0.47] & [0.47,0.498] & & \\
				$$ &  $0.453-0.651$  &  [0.067,0.364] & [0.364,0.394] & [0.394,0.422] & [0.422,0.442] & [0.442,0.472] & [0.148,0.41] & [0.41,0.439] & [0.439,0.468] & [0.468,0.487] & [0.487,0.518] & [0.179,0.483] & [0.483,0.513] & [0.513,0.541] & & \\
				
				$ $ &  $0.651-0.788$  &  [0.096,0.39] & [0.39,0.419] & [0.419,0.448] & [0.448,0.467] & [0.467,0.498] & [0.183,0.435]  &  [0.435,0.465] & [0.465,0.493]&[0.493,0.513]  & [0.513,0.543] & [0.214,0.509] & [0.509,0.539] & [0.539,0.567] & & \\
				$ $ &  $0.788-1$  & [0.118,0.433] & [0.433,0.463] & [0.463,0.491] & [0.491,0.511] & [0.511,0.541] & [0.203,0.479] & [0.479,0.508] & [0.508,0.537] & [0.537,0.556] & [0.556,0.587] & [0.235,0.553] & [0.553,0.582] & [0.582,0.61] & & \\ \hline
				\multirow{5}{*}{0.226-0.348}&$0-0.248$&[0.067,0.323] & [0.323,0.352] & [0.352,0.381] & [0.381,0.4] & [0.4,0.431] & [0.152,0.368] & [0.368,0.398] & [0.398,0.426] & [0.426,0.446] & [0.446,0.476] & [0.183,0.442] & [0.442,0.472] & [0.472,0.5] & & \\
				$ $ &  $0.248-0.453$   &   [0.095,0.366] & [0.366,0.396] & [0.396,0.424] & [0.424,0.444] & [0.444,0.474] & [0.18,0.412] & [0.412,0.441] & [0.441,0.47] & [0.47,0.489] & [0.489,0.52] & [0.211,0.485] & [0.485,0.515] & [0.515,0.543] & & \\
				$ $ & $0.453-0.651$  &  [0.129,0.409] & [0.409,0.439] & [0.439,0.467] & [0.467,0.487] & [0.487,0.517] & [0.215,0.455] & [0.455,0.485] & [0.485,0.513] & [0.513,0.533] & [0.533,0.563] & [0.243,0.529] & [0.529,0.558] & \cellcolor{red!15}[0.558,0.587]& & \\
				$ $ &   $0.651-0.788$  &  [0.164,0.435] & [0.435,0.465] & [0.465,0.493] & [0.493,0.513] & [0.513,0.539] & [0.249,0.481] & [0.481,0.51] & [0.51,0.539] & [0.539,0.558] & [0.558,0.589] & [0.281,0.555] & \cellcolor{red!15}[0.555,0.584] &\cellcolor{red!15} [0.584,0.612]  & & \\
				$ $ & $0.788-1$  & [0.185,0.479] & [0.479,0.508] & [0.508,0.536] & [0.536,0.556] & [0.556,0.586] & [0.27,0.524] & [0.524,0.554] & [0.554,0.582] & [0.582,0.602] & [0.602,0.632] &\cellcolor{red!15} [0.301,0.598] & \cellcolor{red!15}[0.598,0.627] &\cellcolor{red!15} [0.627,0.656] & & \\ \hline
				
				\multirow{5}{*}{0.348-0.496} &  $0-0.248$  &   [0.103,0.377] & [0.377,0.407] & [0.407,0.435] & [0.435,0.455] & [0.455,0.485] & [0.188,0.423] & [0.423,0.452] & [0.452,0.481] & [0.481,0.5] & \cellcolor{red!15}[0.5,0.531] &\cellcolor{red!15} [0.219,0.497] &\cellcolor{red!15} [0.497,0.526] &\cellcolor{red!15} [0.526,0.555]  & &  \\
				$ $ &  $0.248-0.453$   &  [0.131,0.421] & [0.421,0.45] & [0.45,0.479] & [0.479,0.498] & [0.498,0.529] & [0.216,0.466] & [0.466,0.496] & [0.496,0.524] &\cellcolor{red!15} [0.524,0.544] & \cellcolor{red!15}[0.5440.574] & \cellcolor{red!15}[0.248,0.54] & \cellcolor{red!15}[0.54,0.57] &\cellcolor{red!15} [0.57,0.598] & & \\
				$ $ &  $0.453-0.651$  & [0.166,0.464] & [0.464,0.494] & [0.494,0.522] & [0.522,0.542] & [0.542,0.572] & [0.251,0.51] & [0.51,0.539] & \cellcolor{red!15}[0.539,0.568] &\cellcolor{red!15} [0.568,0.587] & \cellcolor{red!15}[0.587,0.612] & \cellcolor{red!15}[0.282,0.584] & \cellcolor{red!15}[0.584,0.613] & \cellcolor{red!15}[0.613,0.642]  & & \\
				$ $ &   $0.651-0.788$  &  [0.2,0.49] & [0.49,0.519] & [0.519,0.548] & [0.548,0.567] & [0.567,0.598] & [0.285,0.535] & \cellcolor{red!15}[0.535,0.565] & \cellcolor{red!15}[0.565,0.593] & \cellcolor{red!15}[0.593,0.613] & \cellcolor{red!15}[0.613,0.643] & \cellcolor{red!15}[0.317,0.609] & \cellcolor{red!15}[0.609,0.639] & \cellcolor{red!15}[0.639,0.667]  & & \\
				$ $ &  $0.788-1$  &  [0.221,0.533] & [0.533,0.563] & [0.563,0.591] & [0.591,0.611] & [0.611,0.641] &\cellcolor{red!15} [0.306,0.579] & \cellcolor{red!15}[0.579,0.608] & \cellcolor{red!15}[0.608,0.637] & \cellcolor{red!15}[0.637,0.656] & \cellcolor{red!15}[0.556,0.687] & \cellcolor{red!15}[0.337,0.653] &\cellcolor{red!15}[0.653,0.682] & \cellcolor{red!15}[0.682,0.711] & & \\ \hline
				
				\multirow{5}{*}{0.496-0.61} &  $0-0.248$ & [0.146,0.42] & [0.42,0.449] & [0.449,0.477] & [0.477,0.497] & \cellcolor{red!15}[0.497,0.528] & \cellcolor{red!15}[0.231,0.465] & \cellcolor{red!15}[0.465,0.495] & \cellcolor{red!15}[0.495,0.523] &\cellcolor{red!15} [0.523,0.543] & \cellcolor{red!15}[0.543,0.573] & \cellcolor{red!15}[0.263,0.539] & \cellcolor{red!15}[0.539,0.568] & \cellcolor{red!15}[0.568,0.597]  & & \\
				$ $ &  $0.248-0.453$   & [0.174,0.463] & [0.463,0.493] & [0.493,0.521] & \cellcolor{red!15}[0.521,0.541] & \cellcolor{red!15}[0.541,0.571] & \cellcolor{red!15}[0.26,0.509] &\cellcolor{red!15}[0.509,0.538] & \cellcolor{red!15}[0.538,0.567] &\cellcolor{red!15} [0.567,0.586] & \cellcolor{red!15}[0.586,0.617] & \cellcolor{red!15}[0.291,0.582] & \cellcolor{red!15}[0.582,0.612] &  \cellcolor{red!40}[0.612,0.64] & & \\
				$ $ &  $0.453-0.651$  & [0.209,0.506] & [0.506,0.536] &\cellcolor{red!15} \cellcolor{red!15}[0.536,0.564] &\cellcolor{red!15} [0.564,0.584] & \cellcolor{red!15}[0.584,0.614] &\cellcolor{red!15} [0.294,0.552] & \cellcolor{red!15}[0.552,0.582] &\cellcolor{red!15} [0.582,0.61] & \cellcolor{red!15}[0.61,0.62] & \cellcolor{red!15}[0.62,0.66]   &  \cellcolor{red!15}[0.326,0.626] & \cellcolor{red!40} [0.626,0.655] &  \cellcolor{red!40}[0.655,0.684] & & \\
				$ $ &   $0.651-0.788$  &  [0.244,0.532] & \cellcolor{red!15}[0.532,0.562] & \cellcolor{red!15}[0.562,0.59] &\cellcolor{red!15} [0.59,0.61]  & \cellcolor{red!15}[0.61,0.64]  &\cellcolor{red!15} [0.329,0.578] & \cellcolor{red!15}[0.578,0.607] & \cellcolor{red!15}[0.607,0.636] &  \cellcolor{red!15}[0.636,0.655] & \cellcolor{red!15} [0.655,0.686] &  \cellcolor{red!40}[0.361,0.651] &  \cellcolor{red!40}[0.651,0.681] &  \cellcolor{red!40}[0.681,0.709] & & \\
				$ $ &  $0.788-1$ &  \cellcolor{red!15}[0.264,0.575] &\cellcolor{red!15} [0.575,0.605] & \cellcolor{red!15}[0.605,0.633] & \cellcolor{red!15}[0.623,0.653] &\cellcolor{red!15} [0.653,0.683] & \cellcolor{red!15}[0.349,0.621] & \cellcolor{red!15}[0.621,0.651] &\cellcolor{red!15}  [0.651,0.679] & \cellcolor{red!15} [0.679,0.699] &  \cellcolor{red!40}[0.699,0.729] &  \cellcolor{red!40}[0.381,0.695] &  \cellcolor{red!40}[0.695,0.724] &  \cellcolor{red!40}[0.724,0.753]  & & \\ \hline
				\multirow{5}{*}{0.61-1.0} &  $0-0.248$  & \cellcolor{red!15}[0.18,0.564] & \cellcolor{red!15}[0.564,0.593] & \cellcolor{red!15}[0.5930.622] & \cellcolor{red!15}[0.622,0.641] &\cellcolor{red!15} [0.641,0.672] & \cellcolor{red!15}[0.265,0.609] &  \cellcolor{red!15}[0.609,0.639] &\cellcolor{red!15} [0.639,0.667] & \cellcolor{red!40} [0.667,0.687] &  \cellcolor{red!40}[0.687,0.717] & \cellcolor{red!40} [0.297,0.683] & \cellcolor{red!40} [0.683,0.713] & \cellcolor{red!40} [0.713,0.741] &&  \\
				$  $ &  $0.248-0.453$   & \cellcolor{red!15}[0.208,0.607] & \cellcolor{red!15}[0.607,0.637] & \cellcolor{red!15}[0.637,0.665] & \cellcolor{red!15}[0.665,0.685] & \cellcolor{red!15}[0.685,0.715] &  \cellcolor{red!15}[0.293,0.653] & \cellcolor{red!15} [0.653,0.682] & \cellcolor{red!40} [0.682,0.711] & \cellcolor{red!40} [0.711,0.73] &  \cellcolor{red!40}[0.73,0.761] &  \cellcolor{red!40}[0.325,0.727] & \cellcolor{red!40} [0.727,0.756] &  \cellcolor{red!40}[0.756,0.785] & & \\
				$  $ &  $0.453-0.651$  &  \cellcolor{red!15}[0.243,0.651] & \cellcolor{red!15}[0.651,0.68] &\cellcolor{red!15} [0.68,0.709] &\cellcolor{red!15} [0.709,0.728] & \cellcolor{red!15}[0.728,0.759] &\cellcolor{red!15} [0.328,0.696] &  \cellcolor{red!40}[0.696,0.726] & \cellcolor{red!40} [0.726,0.754] &  \cellcolor{red!40}[0.754,0.774] & \cellcolor{red!40} [0.774,0.804] &  \cellcolor{red!40}[0.359,0.44] & \cellcolor{red!40}[0.44,0.8] &  \cellcolor{red!40}[0.8,0.828]  & & \\
				
				$  $ &   $0.651-0.788$  & \cellcolor{red!15} [0.278,0.676] & \cellcolor{red!15}[0.676,0.706] & \cellcolor{red!15}[0.706,0.734] & \cellcolor{red!15}[0.734,0.754] & \cellcolor{red!15}[0.754,0.784] &  \cellcolor{red!40}[0.363,0.722] &  \cellcolor{red!40}[0.722,0.751] &  \cellcolor{red!40}[0.751,0.78] &  \cellcolor{red!40}[0.78,0.799] & \cellcolor{red!40} [0.799,0.83] &  \cellcolor{red!40}[0.394,0.796] & \cellcolor{red!40} [0.796,0.825] & \cellcolor{red!40} [0.825,0.854] & & \\
				
				$  $ & $0.788-1$  & \cellcolor{red!15} [0.298,0.72] & \cellcolor{red!15}[0.72,0.749] & \cellcolor{red!15}[0.749,0.778] & \cellcolor{red!15}[0.778,0.797] &  \cellcolor{red!40}[0.797,0.828] & \cellcolor{red!40} [0.383,0.765] &  \cellcolor{red!40}[0.765,0.795] &  \cellcolor{red!40}[0.795,0.823] & \cellcolor{red!40} [0.823,0.843] &  \cellcolor{red!40}[0.843,0.873] &  \cellcolor{red!40}[0.415,0.839] &  \cellcolor{red!40}[0.839,0.869] & \cellcolor{red!60}[0.869,0.897] & & \\  \hline
				\multicolumn{16}{l}{} \\ 
		\end{tabular}}
		\resizebox{\textwidth}{!}{
			\renewcommand{\arraystretch}{1.8}
			\begin{tabular}{|c|c|cc|ccccc|ccccc|cccc}
				\multicolumn{16}{l}{\large{continue from previous table}} \\ \cline{1-14}
				& VCR & \multicolumn{2}{c|}{0.42-0.604}& \multicolumn{5}{c|}{0.604-0.759}& \multicolumn{5}{c|}{0.759-1.0} &\makebox[0.02\textwidth][c]{}  & & \makebox[0.04\textwidth][c]{}&\\ \cline{1-14}
				
				ECR &\diagbox{\makebox[0.035\textwidth][c]{VHIR}}{\makebox[0.035\textwidth][c]{VCCR}}  & 0.639-0.773&0.773-1& 0-0.184 &  0.184-0.412 & 0.412-0.639&  0.639-0.773&0.773-1& 0-0.184&  0.184-0.412  & 0.412-0.639 &  0.639-0.773&0.773-1&  && &\multirow{3}{*}{{\color{red}Dangerous}} \\ \cline{1-14}
				
				\multirow{5}{*}{ 0-0.226} &  $0-0.248$&[0.455,0.474] & [0.474,0.505] & [0.168,0.459] & [0.459,0.488] & [0.488,0.517] & [0.517,0.536] & \cellcolor{red!15}[0.536,0.567] & \cellcolor{red!15}[0.211,0.556] &\cellcolor{red!15} [0.556,0.585] & \cellcolor{red!15}[0.585,0.614] &\cellcolor{red!15} [0.614.0.633] &\cellcolor{red!15} [0.633,0.664] &  & \multirow{5}{*}{0.903-1} & \cellcolor{red!100}\multirow{5}{*}{Dangerous}& \multirow{5}{*}{Extremely Serious}\\
				$ $ &  $0.248-0.453$& [0.498,0.518] & [0.518,0.548] & [0.196,0.502] & [0.502,0.532] & [0.532,0.56] &\cellcolor{red!15} [0.56,0.58] &\cellcolor{red!15}[0.58,0.61] &     \cellcolor{red!15}[0.239,0.599] &\cellcolor{red!15} [0.599,0.629] & \cellcolor{red!15}[0.629,0.657] & \cellcolor{red!15}[0.657,0.677] & \cellcolor{red!15}[0.677,0.707]  &  &&\cellcolor{red!100} & \\
				$$ &  $0.453-0.651$ & [0.541,0.561] & [0.561,0.591] & [0.23,0.546] & [0.546,0.575] & \cellcolor{red!15}[0.575,0.604] & \cellcolor{red!15}[0.604,0.623] & \cellcolor{red!15}[0.623,0.654] & \cellcolor{red!15}[0.274,0.643] & \cellcolor{red!15}[0.643,0.672] & \cellcolor{red!15}[0.672,0.7] & \cellcolor{red!15}[0.7,0.72] &     \cellcolor{red!15}[0.72,0.75] &  &&  \cellcolor{red!100}&\\
				$ $ &  $0.651-0.788$ & [0.567,0.587] & [0.587,0.617] & [0.265,0.571] & \cellcolor{red!15}[0.571,0.601] &\cellcolor{red!15} [0.601,0.629] & \cellcolor{red!15}[0.629,0.649] &\cellcolor{red!15} [0.649,0.679] & \cellcolor{red!15}[0.308,0.668] & \cellcolor{red!15}[0.668,0.698] & \cellcolor{red!15}[0.698,0.726] & \cellcolor{red!15}[0.726,0.746] & \cellcolor{red!15}[0.746,0.776] &  &  &\cellcolor{red!100}&\\
				$ $ &  $0.788-1$ & [0.61,0.63] &     [0.63,0.66] &\cellcolor{red!15}[0.286,0.615] & \cellcolor{red!15}[0.615,0.644] & \cellcolor{red!15}[0.644,0.673] &\cellcolor{red!15} [0.673,0.692] & \cellcolor{red!15}[0.692,0.723] & \cellcolor{red!15}[0.329,0.712] & \cellcolor{red!15}[0.712,0.741] & \cellcolor{red!15}[0.741,0.769] &\cellcolor{red!15} [0.769,0.789] &  \cellcolor{red!40}[0.789,0.82]  &  & &\cellcolor{red!100} &\\ \cline{1-14}
				
				\multirow{5}{*}{0.226-0.348}  &  $0-0.248$ & [0.5,0.52] &     \cellcolor{red!15}[0.52,0.55]  & \cellcolor{red!15}[0.234,0.504] & \cellcolor{red!15}[0.504,0.534] & \cellcolor{red!15}[0.534,0.562] & \cellcolor{red!15}[0.562,0.582] & \cellcolor{red!15}[0.582,0.612] & \cellcolor{red!15}[0.277,0.601] &\cellcolor{red!15} [0.601,0.631] & \cellcolor{red!15}[0.631,0.659] &  \cellcolor{red!40}[0.659,0.679] & \cellcolor{red!40} [0.679,0.709] && \multirow{5}{*}{0.758-0.903}  & \cellcolor{red!60}\multirow{5}{*}{Dangerous}&\multirow{5}{*}{Serious}\\
				
				$ $ &  $0.248-0.453$ & \cellcolor{red!15} [0.543,0.563] &\cellcolor{red!15} [0.563,0.593] & \cellcolor{red!15}[0.262,0.548] & \cellcolor{red!15}[0.548,0.577] &\cellcolor{red!15} [0.577,0.606] & \cellcolor{red!15}[0.606,0.625] &\cellcolor{red!15} [0.625,0.656] & \cellcolor{red!15}[0.305,0.644] & \cellcolor{red!15}[0.644,0.674] &  \cellcolor{red!40}[0.674,0.702] & \cellcolor{red!40} [0.702,0.722] &  \cellcolor{red!40}[0.722,0.752] &  & &\cellcolor{red!60}& \\
				
				$ $ & $0.453-0.651$  &\cellcolor{red!15} [0.587,0.606] & \cellcolor{red!15}[0.606,0.637] & \cellcolor{red!15}[0.297,0.591] & \cellcolor{red!15}[0.591,0.621] & \cellcolor{red!15}[0.621,0.649] & \cellcolor{red!15}[0.649,0.669] &\cellcolor{red!15} [0.669,0.699] & \cellcolor{red!15}[0.34,0.688] &  \cellcolor{red!40}[0.688,0.717] & \cellcolor{red!40}[0.717,0.746] &  \cellcolor{red!40}[0.746,0.765] &  \cellcolor{red!40}[0.765,0.796]   & & &\cellcolor{red!60}& \\
				
				$ $ &   $0.651-0.788$ & \cellcolor{red!15}[0.612,0.632] & \cellcolor{red!15}[0.632,0.662] & \cellcolor{red!15}[0.332,0.617] & \cellcolor{red!15}[0.617,0.646] &\cellcolor{red!15} [0.646,0.675] & \cellcolor{red!15}[0.675,0.694] &\cellcolor{red!15} [0.694,0.725] &  \cellcolor{red!40}[0.375,0.714] & \cellcolor{red!40} [0.714,0.743] & \cellcolor{red!40} [0.743,0.771] &  \cellcolor{red!40}[0.771,0.791] &  \cellcolor{red!40}[0.791,0.822]&  &&\cellcolor{red!60} &\\
				
				$ $ & $0.788-1$& \cellcolor{red!15}[0.656,0.675] & \cellcolor{red!15}[0.675,0.706] & \cellcolor{red!15}[0.352,0.66] & \cellcolor{red!15}[0.66,0.69] & \cellcolor{red!15}[0.69,0.718] & \cellcolor{red!15}[0.718,0.738] &  \cellcolor{red!40}[0.738,0.768] &  \cellcolor{red!40}[0.395,0.757] & \cellcolor{red!40} [0.757,0.786] & \cellcolor{red!40} [0.786,0.815] & \cellcolor{red!40} [0.815,0.835] &  \cellcolor{red!40}[0.835,0.865] &  &&\cellcolor{red!60} &\multirow{2}{*}{{\color{orange}Critical}}\\ \cline{1-14}
				
				\multirow{5}{*}{0.348-0.496} &  $0-0.248$  & \cellcolor{red!15} [0.555,0.574] & \cellcolor{red!15}[0.574,0.605] &\cellcolor{red!15} \cellcolor{red!15}[0.27,0.559] & \cellcolor{red!15}[0.559,0.588] & \cellcolor{red!15}[0.588,0.617] &  \cellcolor{red!40}[0.617,0.636] & \cellcolor{red!40} [0.636,0.667] & \cellcolor{red!40} [0.313,0.656] &  \cellcolor{red!40}[0.656,0.685] &  \cellcolor{red!40}[0.685,0.714] &  \cellcolor{red!40}[0.714,0.733] & \cellcolor{red!40} [0.733,0.764] &  & \multirow{5}{*}{0.506-0.758} &\cellcolor{red!40}&\multirow{5}{*}{Slightly Serious}\\
				
				$ $ &  $0.248-0.453$  & \cellcolor{red!15}[0.598,0.618] & \cellcolor{red!15}[0.618,0.648] &\cellcolor{red!15} [0.299,0.602] & \cellcolor{red!15}[0.602,0.632] &  \cellcolor{red!40}[0.632,0.66] &  \cellcolor{red!40}[0.66,0.68] & \cellcolor{red!40}[0.68,0.71]     &  \cellcolor{red!40}[0.342,0.699] &  \cellcolor{red!40}[0.699,0.729] &  \cellcolor{red!40}[0.729,0.757] & \cellcolor{red!40} [0.757,0.777] &  \cellcolor{red!40}[0.777,0.807]&  & &\cellcolor{red!40}& \\
				
				$ $ &  $0.453-0.651$ &  \cellcolor{red!15}[0.642,0.661] & \cellcolor{red!15}[0.661,0.692] & \cellcolor{red!15}[0.333,0.646] & \cellcolor{red!40} [0.646,0.675] &  \cellcolor{red!40}[0.675,0.704] & \cellcolor{red!40} [0.704,0.723] & \cellcolor{red!40} [0.723,0.754] &  \cellcolor{red!40}[0.376,0.743] &  \cellcolor{red!40}[0.743,0.772] &  \cellcolor{red!40}[0.772,0.801] &  \cellcolor{red!40}[0.801,0.82] &\cellcolor{red!60} [0.82,0.851] & & &\cellcolor{red!40} & \\
				
				$ $ &   $0.651-0.788$  & \cellcolor{red!15}[0.667,0.687] & \cellcolor{red!15}[0.687,0.717] &  \cellcolor{red!40}[0.368,0.671] & \cellcolor{red!40} [0.671,0.701] & \cellcolor{red!40} [0.7010.729] &  \cellcolor{red!40}[0.729,0.749] & \cellcolor{red!40} [0.749,0.779] &  \cellcolor{red!40}[0.411,0.768] &  \cellcolor{red!40}[0.768,0.798] & \cellcolor{red!40} [0.789,0.826] & \cellcolor{red!60}[0.826,0.846] & \cellcolor{red!60}[0.846,0.876]  & &&\cellcolor{red!40}&\\
				
				$ $ &  $0.788-1$ &\cellcolor{red!15}[0.711,0.73] &  \cellcolor{red!40}[0.73,0.761] & \cellcolor{red!40} [0.389,0.715] &  \cellcolor{red!40}[0.715,0.744] & \cellcolor{red!40} [0.744,0.773] &  \cellcolor{red!40}[0.773,0.792] &  \cellcolor{red!40}[0.792,0.823] &  \cellcolor{red!40}[0.432,0.812] &  \cellcolor{red!40}[0.812,0.841] & \cellcolor{red!60}[0.841,0.87] &\cellcolor{red!60} [0.87,0.889] & \cellcolor{red!60}[0.889,0.92]& & & \cellcolor{red!40}&\multirow{2}{*}{{\color{blue}Steady}}\ \\
				\cline{1-14}
				
				\multirow{5}{*}{0.496-0.61} &  $0-0.248$ &  \cellcolor{red!40}[0.597,0.617] &  \cellcolor{red!40}[0.617,0.647] &  \cellcolor{red!40}[0.314,0.601] &  \cellcolor{red!40}[0.601,0.63]1 &  \cellcolor{red!40}[0.63,0.659] &  \cellcolor{red!40}[0.659,0.679] &  \cellcolor{red!40}[0.679,0.709] &  \cellcolor{red!40}[0.357,0.698] & \cellcolor{red!60}[0.698,0.727] &\cellcolor{red!60} [0.727,0.756] & \cellcolor{red!60}[0.756,0.776] & \cellcolor{red!60}[0.766,0.806] &&\multirow{5}{*}{0.264-0.504}&\multirow{5}{*}{danger}\cellcolor{red!15} & \multirow{5}{*}{Slight}\\
				
				$ $ &  $0.248-0.453$ &  \cellcolor{red!40}[0.64,0.66] & \cellcolor{red!40}[0.66,0.69] &  \cellcolor{red!40}[0.342,0.645] &  \cellcolor{red!40}[0.645,0.674] &  \cellcolor{red!40}[0.674,0.702] &  \cellcolor{red!40}[0.702,0.722] &  \cellcolor{red!40}[0.722,0.753] & \cellcolor{red!60}[0.382,0.741] & \cellcolor{red!60}[0.941,0.771] &\cellcolor{red!60} [0.771,0.799] &\cellcolor{red!60} [0.799,0.819] &\cellcolor{red!60} [0.819,0.849] && & \cellcolor{red!15}&\\
				
				$ $ &  $0.453-0.651$ &   \cellcolor{red!40}[0.684,0.703] &  \cellcolor{red!40}[0.703,0.734] &  \cellcolor{red!40}[0.377,0.688] &  \cellcolor{red!40}[0.688,0.717] & \cellcolor{red!40} [0.717,0.746] &  \cellcolor{red!40}[0.746,0.766] &\cellcolor{red!60} [0.766,0.796] &\cellcolor{red!60} [0.42,0.785] & \cellcolor{red!60}[0.785,0.814] &\cellcolor{red!60} [0.814,0.843] &\cellcolor{red!60} [0.843,0.862] & \cellcolor{red!60}[0.862,0.893] & & &\cellcolor{red!15}& \\
				
				$ $ &   $0.651-0.788$ &  \cellcolor{red!40}[0.709,0.729] &  \cellcolor{red!40}[0.729,0.759] &  \cellcolor{red!40}[0.412,0.714] &  \cellcolor{red!40}[0.714,0.743] &  \cellcolor{red!40}[0.743,0.772] & \cellcolor{red!60}[0.772,0.791] &\cellcolor{red!60} [0.791,0.822] & \cellcolor{red!60}[0.455,0.81] &\cellcolor{red!60} [0.81,0.84]   & \cellcolor{red!60}[0.84,0.868] & \cellcolor{red!60}[0.868,0.888] & \cellcolor{red!60}[0.888,0.918] &&   &\cellcolor{red!15} &  \\
				
				$ $ &  $0.788-1$ &  \cellcolor{red!40}[0.753,0.772] &  \cellcolor{red!40}[0.772,0.803] &  \cellcolor{red!40}[0.432,0.757] &  \cellcolor{red!40}[0.757,0.786] & \cellcolor{red!60}[0.786,0.815] & \cellcolor{red!60}[0.815,0.835] & \cellcolor{red!60}[0.835,0.865] &\cellcolor{red!60} [0.475,0.854] & \cellcolor{red!60}[0.854,0.883] & \cellcolor{red!60}[0.883,0.912] & \cellcolor{red!60}[0.912,0.931] & \cellcolor{red!100}[0.931,0.962]& & & \cellcolor{red!15}& \\ \cline{1-14}
				
				\multirow{5}{*}{0.61-1.0} &  $0-0.248$  &  \cellcolor{red!40}[0.741,0.761] &  \cellcolor{red!40}[0.761,0.791] &  \cellcolor{red!40}[0.348,0.745] & \cellcolor{red!60}[0.745,0.775] &\cellcolor{red!60} [0.775,0.803] &\cellcolor{red!60} [0.803,0.823] & \cellcolor{red!60}[0.823,0.853] &\cellcolor{red!60} [0.391,0.842] & \cellcolor{red!60}[0.842,0.872] & \cellcolor{red!60}[0.872,0.9] &\cellcolor{red!100}[0.90.92] &\cellcolor{red!100}[0.92,0.95]  &   & \multirow{5}{*}{0-0.264} && \multirow{5}{*}{Normal}  \\
				
				$  $ &  $0.248-0.453$ &  \cellcolor{red!40} [0.785,0.804] &  \cellcolor{red!40}[0.804,0.835] & \cellcolor{red!60}[0.376,0.789] & \cellcolor{red!60}[0.789,0.818] & \cellcolor{red!60}[0.818,0.847] & \cellcolor{red!60}[0.847,0.866] & \cellcolor{red!60}[0.866,0.897] &\cellcolor{red!60} [0.419,0.886] & \cellcolor{red!60}[0.886,0.915] & \cellcolor{red!100}[0.915,0.944] & \cellcolor{red!100}[0.944,0.963] &\cellcolor{red!100} [0.963,0.994]  & && &\\
				
				$  $ &  $0.453-0.651$ & \cellcolor{red!40}[0.828,0.848] &\cellcolor{red!60} [0.848,0.878] & \cellcolor{red!60}[0.411,0.832] & \cellcolor{red!60}[0.832,0.862] & \cellcolor{red!60}[0.862,0.89] &\cellcolor{red!60} [0.89,0.91] &     \cellcolor{red!60}[0.91,0.94]  & \cellcolor{red!60}[0.454,0.93] & \cellcolor{red!100}[0.93,0.959] & \cellcolor{red!100}[0.959,0.987] & \cellcolor{red!100}1 &\cellcolor{red!100}1 &  & & &\\
				
				$  $ & $0.651-0.788$&\cellcolor{red!60}[0.854,0.873] & \cellcolor{red!60}[0.873,0.904] & \cellcolor{red!60}[0.445,0.858] &\cellcolor{red!60} [0.858,0.887] &\cellcolor{red!60} [0.887,0.916] &\cellcolor{red!60} [0.916,0.935] &\cellcolor{red!60} [0.935,0.966] & \cellcolor{red!100}[0.488,0.955] & \cellcolor{red!100}[0.955,0.984] &\cellcolor{red!100}[0.984,1] &\cellcolor{red!100}1 & \cellcolor{red!100}1  & & &&\\
				
				$$&$0.788-1$&\cellcolor{red!60}[0.897,0.912] & \cellcolor{red!60}[0.912,0.947] &\cellcolor{red!60} [0.466,0.901] & \cellcolor{red!60}[0.901,0.931] & \cellcolor{red!60}[0.931,0.959] & \cellcolor{red!60}[0.959,0.979] & \cellcolor{red!100} [0.979,1] &\cellcolor{red!100}  [0.509,0.998] & \cellcolor{red!100} [0.998,1] & \cellcolor{red!100} 1 & \cellcolor{red!100} 1 & \cellcolor{red!100} 1  &  & &&
				\\ \cline{1-14}
				
			\end{tabular}
			
	}}
	\label{table:13}
\end{table*}
\subsection{Analysis Of Evaluation Results}
The safety of vehicle intelligent connected systems on the OpenPilot platform is then affected by the metrics we suggested in Section 3 in this section.


\begin{figure}[!]
	\centering
	\includegraphics[width=0.7\linewidth]{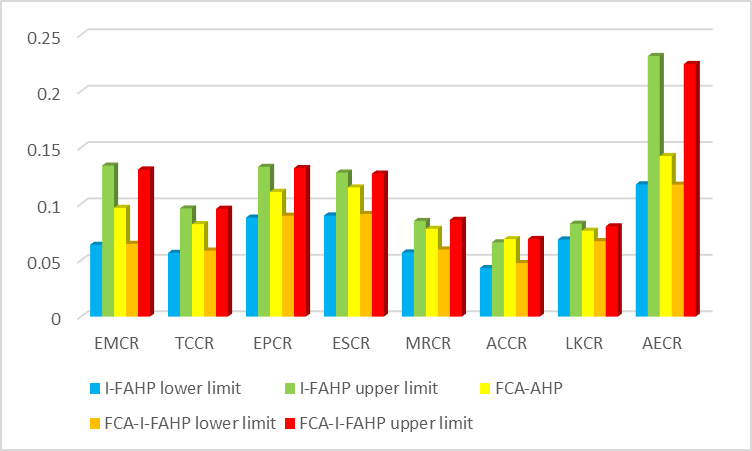}
	\captionsetup{font={scriptsize}}
	\caption{\scriptsize{Comparison of ECR risk assessment value of OpenPilot}}
	\label{fig:ecr2}
	
	\centering
	\includegraphics[width=0.7\linewidth]{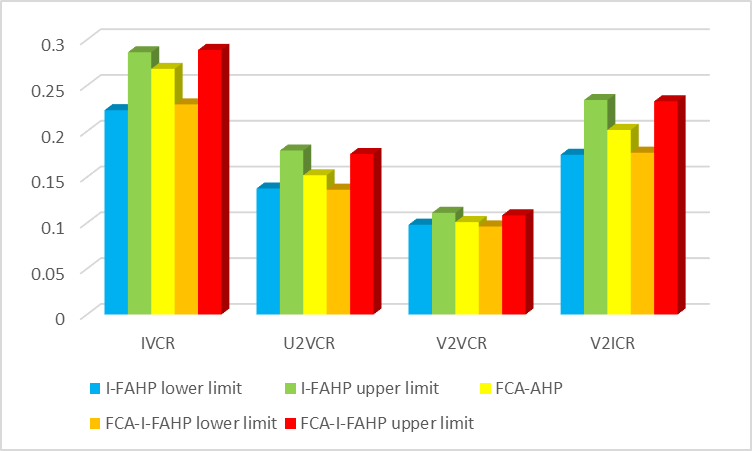}
	\caption{Comparison of VCR risk assessment value of OpenPilot}
	\label{fig:vcr2}
	
	\centering
	\includegraphics[width=0.7\linewidth]{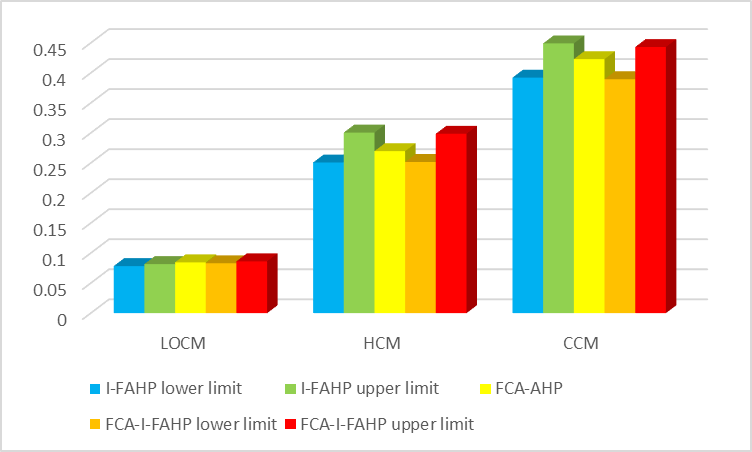}
	\caption{Comparison of VCCR risk assessment value of OpenPilot }
	\label{fig:vccr2}
	
	\centering
	\includegraphics[width=0.7\linewidth]{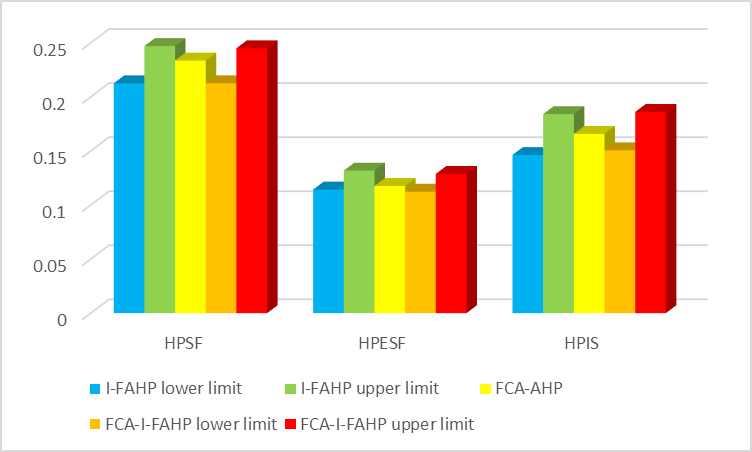}
	\caption{Comparison of VHIR risk assessment value of OpenPilot }
	\label{fig:vhsr2}
\end{figure}

\begin{table*}[!]
	\centering	
	\renewcommand{\arraystretch}{2.3}
	\caption{Data Analysis of Risk Influencing Factors of OpenPilot System}
	\begin{threeparttable}
		\resizebox{\textwidth}{!}{
			\begin{tabular}{p{3.3cm}<{\centering}|p{5.6cm}<{\centering}| p{9cm}<{\centering}} \hline
				\textbf{ System Risk Factors}   & \makebox[0.02\textwidth][c]{\textbf{ Risk Assessment Value} }  &\makebox[0.05\textwidth][c]{\textbf{Risk Assessment Calculation Detail}}      \\ \hline 
				
				ECU Coupling Risk (ECR) &ECR(OpenPilot)=[0.57293920,0.94365520] &
				The OpenPilot controls the major components of the vehicle, including ECM$^1$, BCM$^2$, SSM$^3$, SCU$^4$, PCM$^5$, TCM$^6$, TCU$^7$, EPS$^8$        \\ \hline
				
				Vehicle Communication Risk(VCR) & VCR(OpenPilot)=[0.63848454,0.80522362] &The OpenPilot system controls the vehicle's communication modules, including In-Vehicle, U2V, V2V and V2I communications   \\ \hline
				
				Vehicle Code Complexity Risk (VCCR) &VCCR(OpenPilot)=[0.72454854,0.82819294] &The OpenPilot has a total of 1684 files, including 234464 lines of code, 31925 blank lines and 34723 comment lines\\ \hline
				
				Risk of Vehicle History Security Issues(VHIR) &VHIR(OpenPilot)=[0.47562178,0.56006950] & The OpenPilot has reported 175 bugs since 2018(13 in 2018,51 in 2019,61 in 2020,31 in 2021, and 19 in 2022)   \\ \hline
				
			\end{tabular}
		}
	\end{threeparttable}
	\begin{tablenotes}
		\footnotesize
		\item{$^1$}Engine control module, {$^2$}Brake control module, {$^3$}Safety system module, {$^4$}Seat control unit, {$^5$}Powertrain control module
		\item{$^6$}Transmission control module, {$^7$}Telematics control unit, {$^8$}Electric power steering
		control module
	\end{tablenotes}
	\label{table:15}
\end{table*}

\subsubsection{OpenPilot Platform Experiment}

By looking at the kind of CAN messages being received in the automated driving system, we were able to determine the vehicle's ECU coupling risk information based on equation (1). 
Since the vehicle's automatic driving function only uses two types of communication—V2V for receiving information about the locations of nearby vehicles and in-vehicle communication for connecting with sensors, radars, and ECUs—we can only estimate the communication risk of the vehicle's intelligent connected system using the equations (2) (3) (4) (5).
The number of source code lines (LOCM), operators and operands (HCM), and cyclomatic complexity (CCM) of the vehicle intelligent connected system were then determined using equations (6), (7), and (8), and the risk associated with code complexity for the vehicle was then determined. 
Equations (9), (10) and (11) are used to compute the historical risk of the vehicle's functional safety, expected functional safety, and information security. 
This information is then used to determine the historical risk of the vehicle's safety.

Then, through the vehicle impact index data calculated by the above equation, in order to evaluate the risk of the vehicle intelligent connected system more conveniently and accurately, we then use the equation (14) to normalize the above data, so we can get the Table \ref{table:15}.

Then, based on the risk index data of various vehicle intelligent connected systems on the OpenPilot platform listed in Table \ref{table:15} above, combined with the objective function equation (13) given in this paper, we can calculate the risk assessment value of the OpenPilot system as [0.514567,0.911115] , and this value is exactly the serious risk level in the defined security risk evaluation interval. This is also in line with the public OpenPilot system ASIL level D=S3 + E4 + C3, where level D is the highest level of risk rating in the ISO 26262 standard, which is also consistent with the evaluation results of this paper, which also proves our evaluation the results are solid.

Among them, Fig. \ref{fig:ecr2}, \ref{fig:vcr2}, \ref{fig:vccr2}, and \ref{fig:vhsr2} show that the modified weights obtained from I-FAHP and FCA-I-FAHP ECU coupling risk, communication risk, code complexity risk, vehicle history safety event risk and vehicle comprehensive risk of the vehicle intelligent connected system. 
And based on the aforementioned we can also see in each picture that the findings of the risk assessment for the FCA-I-FAHP method and the I-FAHP method for the vehicle intelligent connected system are comparable. 
However, it can be seen from comparing the model's results using the AHP and FCA-I-FAHP approaches that the FCA-I-FAHP method can more precisely and comprehensively determine the current risk level of the vehicle intelligent connected system. 
The findings of the I-FAHP and FCA-I-FAHP approaches may be compared, and this comparison can also demonstrate that the I-FAHP and FCA approach can be used to determine risk levels for higher risk vehicle intelligent connected systems.

As shown in Fig. \ref{fig:ecr2}, \ref{fig:vcr2}, \ref{fig:vccr2}, after the evaluation of the above-mentioned model FCA-I-FAHP method, the ECU coupling risk, communication risk, and code complexity risk of the vehicle are at the severity level; while the results in Fig. \ref{fig:vhsr2} show that the vehicle’s the risk level of historical security incidents is at the minor risk level. The comparison of the experimental results confirms that the ECU coupling risk, communication risk and code complexity risk of the vehicle exacerbate the risk of the OpenPilot vehicle intelligent connected system.

\subsubsection{Improvements For The OpenPilot Platform}

We may determine that the OpenPilot system is in a risky state by the computation of the aforementioned model technique, so it is required to make improvements to the OpenPilot system until it reaches a steady state. 

As a result, we compare the number of security vulnerabilities discovered in each file on the OpenPilot system in the past in order to assess the efficacy of our model approach improvement methods. 
On the OpenPilot platform, there are a total of 1684 files and 175 error files among them as of December 2022. After reviewing the reported bugs, we linked 64 of the files with the issues that had been fixed above to the system files.

Then, using our above-mentioned vehicle intelligent connected system risk assessment model, compared with the AVSDA\cite{moukahal2022avsda} vulnerability measurement tool, code complexity and loss measurement tool \cite{ShiMW2010} and coding complexity, code coupling and cohesion measurement tool \cite{ChoZ2011}, through the binary classification method , identify the true positive(TP), false positive(FP), true negative(TN), and false negative(FN) files in the experiment, and measure the accuracy, precision, and recall of each model.

The above results are shown in Table \ref{table:VSRQ model}.
Through this table, we can know that the proposed VSRQ model is superior to other methods in terms of accuracy, precision and recall.
The recall rate for our suggested model among them was 73.43{\%}, showing that it is capable of quickly locating risk files in vehicle intelligent connected systems. 
The VSRQ model we suggested also has an accuracy rate as high as 94.36{\%}, suggesting that the identification of vehicle systems and risk assessment are both fairly reliable. Because we use a more thorough two-stage indicator evaluation in the model, the VSRQ model is also superior to several other methods in terms of precision and has a precision of 37.60{\%}. 

\begin{figure}[!]
	\centering
	\includegraphics[width=0.8\linewidth]{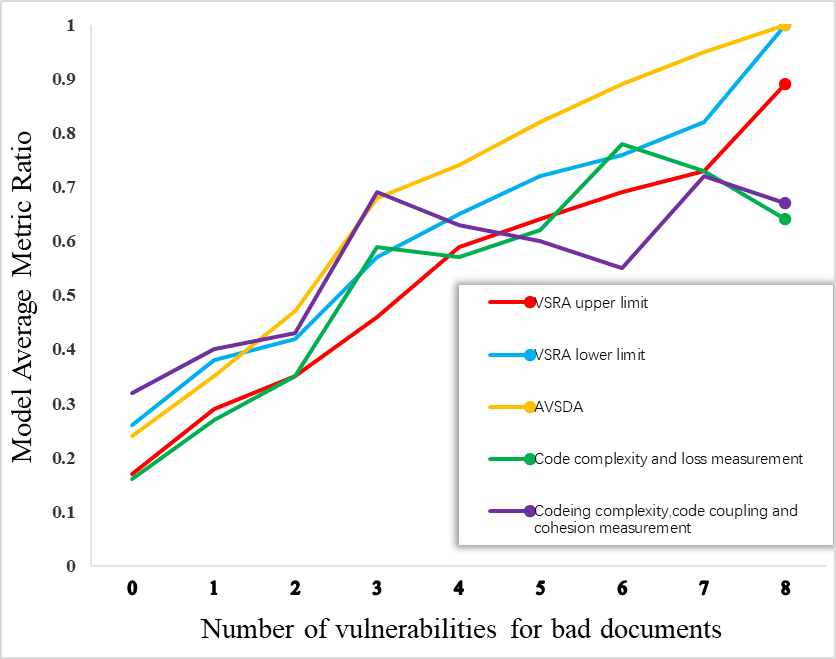}
	\caption{Idicator ratios for each risk assessment model}
	\label{fig:average_indicator}	
\end{figure}

Finally, we also examine the relationship between the average metric ratio and the quantity of errors recorded in the file, as shown in Fig. \ref{fig:average_indicator}, in order to further investigate the performance of the VSRQ. 
For instance, the VSRQ classifies a file with 8 reported vulnerabilities as having an extremely significant risk level and assigns it a ratio of 1.
As seen in Fig. \ref{fig:average_indicator}, the risk rating coefficient VSR determined by the VSRQ is related to the number of bugs in the wrong files in the OpenPilot system, and the greater the value of the VSR risk coefficient, the more bugs there are in the bug files. When compared to the other three, we can see that the AVSDA\cite{moukahal2022avsda} model's recognition accuracy is insufficient, there is no intervalization, and the other two groups of metrics' \cite{ShiMW2010},\cite{ChoZ2011} values are chosen at random.

\begin{table}[h]
	\centering	
	\renewcommand{\arraystretch}{1.2}
	\caption{Comparison between VSRQ and other models}
	\begin{threeparttable}
		\begin{tabular}{c|cccc} \hline
			&\textbf{\textbf{VSRQ} }   &\textbf{\textit{AVSDA\cite{moukahal2022avsda}}}    &   \textbf{\textit{CCL$^a$ \cite{ShiMW2010}}}  & \textbf{\textit{ CCC$^b$\cite{ChoZ2011} }}   \\ \hline
			TN  &  1542    &  1517     &  1528& 1483\\ 
			TP    &47   &  41   &  35 &26\\
			FN &17  & 23 & 29&38\\ 
			FP  &78    &  103   &  92&137 \\ 
			Accuracy   &94.36$\%$  &  92.52$\%$   &  92.81$\%$ &89.61$\%$\\ 
			Precision  &37.60$\%$   &  28.47$\%$  &  27.56$\%$ &15.95$\%$\\  
			Recall &73.43$\%$   & 64.06$\%$  & 54.69$\%$ &40.63$\%$\\ \hline
			
		\end{tabular}
	\end{threeparttable}
	\begin{tablenotes}
		\footnotesize
		\item{$^a$} Code complexity and loss measurement
		\item{$^b$} Coding complexity, code coupling and cohesion measurement
	\end{tablenotes}
	\label{table:VSRQ model}
\end{table}
\subsubsection{Performance Comparison of VSRQ Model Method}

Then we compared it with other three methods from the perspective of analysis process and analysis results, including AVSDA\cite{moukahal2022avsda}, code complexity and churn measurement tool \cite{ShiMW2010} and coding complexity, code coupling and cohesion measurement tool \cite{ChoZ2011}.

According to the experiment's process analysis, we can infer that the AVSDA's \cite{moukahal2022avsda} primary indicators are overly simplistic, and its weighting of the indicators is too arbitrary, which frequently prevents the identification of vulnerabilities. In contrast, the code complexity and loss measurement tool \cite{ShiMW2010}and the coding complexity, code coupling, and cohesion measurement tool \cite{ChoZ2011} only examine the code complexity of the vehicle intelligent connected system separately.
While the VSRQ proposed in this paper not only takes into account the multi-indicator influencing factors that pose risks to the vehicle system, but also introduces the FCA method to examine the relationship between the influencing factors of the vehicle system and the ISO 26262 \cite{PalWH2011}evaluation standard, the consideration is incredibly simple and lacks the rationality of the analysis.
The FCA-I-FAHP approach can be used to identify greater risk vehicle software integrated system risk levels as well as reduce the subjectivity of the AHP method.

We compared the risk analysis outcomes from the four approaches mentioned above for the experimental findings. 
We may infer from Table \ref{table:15} and Fig. \ref{fig:average_indicator} that the proposed VSRQ model outperforms alternative approaches in terms of accuracy, precision, and recall.

\section{Conclusion}
\label{sec:conclusion}

In this research, we present a method for quantitatively assessing the security risk of vehicle intelligent connected systems called the risk assessment of vehicle intelligent connected system (VSRQ).
There are two stages to the VSRQ approach.
We create safety measurement equations for its sub-influencing factors during the analysis stage of vehicle vulnerable parts using the risk assessment categories of ECU coupling, communication risk, code complexity risk, and vehicle history safety event risk.
We used the interval FAHP approach for risk assessment, weighing the relative relevance of susceptible vehicle components in relation to the severity of vehicle damage. 
We then used the equation VSR to weigh the likelihood of safety concerns in software-integrated vehicle system.
We create the vehicle software system security risk matrix integration, as shown in Fig.\ref{table:13}.

Then, we identified three states of the vehicle intelligent connected system—stable, critical, and dangerous—and proposed corrective safety actions for each.
The findings demonstrate that the evaluation approach can pinpoint vulnerabilities with a 94.36{\%} accuracy rate and a 73.43{\%} recall rate.
We're eager to investigate more situations and try VSRQ on different platforms.
In addition, our vehicle intelligent connected solutions can be used to assess the risk associated with certain vehicle parts, such as the engine system.

\end{document}